\documentclass[10pt,twocolumn,letterpaper]{article}
\usepackage{cvpr}              %
\usepackage{times}
\usepackage{epsfig}
\usepackage{graphicx}
\usepackage{amsmath}
\usepackage{amssymb}
\usepackage{enumitem}
\usepackage{hyperref}

\AtBeginDocument{%
  }

\usepackage{comment}
\usepackage{hhline}
\usepackage{multirow}
\usepackage{xspace}

\usepackage[acronym, nonumberlist]{glossaries}
\newacronym{ai}{AI}{Artificial Intelligence}
\newacronym{ml}{ML}{Machine Learning}
\newacronym{nlp}{NLP}{Natural Language Processing}
\newacronym{nli}{NLI}{Natural Language Inference}
\newacronym{mnli}{MNLI}{Multi-Genre Natural Language Inference}
\newacronym{vlm}{VLM}{Vision-Language Model}
\newacronym{vqa}{VQA}{Visual Question Answering}
\newacronym{se}{SE}{Semantic Entropy}
\newacronym{vase}{VASE}{Vision-Amplified Semantic Entropy}
\newacronym{hedge}{HEDGE}{Hallucination Estimation via Dense Geometric Entropy}
\newacronym{uq}{UQ}{Uncertainty Quantification}
\newacronym{auc}{AUC}{Area Under the Curve}
\newacronym{roc}{ROC}{Receiver Operating Characteristic}
\newacronym{rocauc}{ROC-AUC}{Receiver Operating Characteristic Area Under the Curve}
\newacronym{clip}{CLIP}{Contrastive Language-Image Pre-training}
\newacronym{llm}{LLM}{Large Language Model}

\newacronym{cnn}{CNN}{Convolutional Neural Network}
\newacronym{vit}{ViT}{Vision Transformer}
\newacronym{lstm}{LSTM}{Long Short-Term Memory}
\newacronym{rnn}{RNN}{Recurrent Neural Network}
\newacronym{nn}{NN}{Neural Network}
\newacronym{cv}{CV}{Computer Vision}

\glsdisablehyper

\usepackage{listings}
\usepackage{array}
\makeglossaries

\makeglossaries

\begin{document}

\title{HEDGE: Hallucination Estimation via Dense Geometric Entropy \\
for VQA with Vision--Language Models}

\author{
Sushant Gautam$^{1,2}$ \qquad
Michael A.~Riegler$^{3}$ \qquad
P\aa l Halvorsen$^{1,2}$ \\
{\tt\small sushant@simula.no} \qquad
{\tt\small michael@simula.no} \qquad
{\tt\small paalh@simula.no} \\
$^{1}$Simula Metropolitan Center for Digital Engineering (SimulaMet), Oslo, Norway \\
$^{2}$Oslo Metropolitan University (OsloMet), Oslo, Norway \\
$^{3}$Simula Research Laboratory, Oslo, Norway
}

\maketitle
\thispagestyle{empty}
\begin{abstract}
Vision--language models (VLMs) enable open-ended visual question answering but remain prone to hallucinations. We present HEDGE, a unified framework for hallucination detection that combines controlled visual perturbations, semantic clustering, and robust uncertainty metrics. HEDGE integrates sampling, distortion synthesis, clustering (entailment- and embedding-based), and metric computation into a reproducible pipeline applicable across multimodal architectures.

Evaluations on VQA-RAD and KvasirVQA-x1 with three representative VLMs (LLaVA-Med, Med-Gemma, Qwen2.5-VL) reveal clear architecture- and prompt-dependent trends. Hallucination detectability is highest for unified-fusion models with dense visual tokenization (Qwen2.5-VL) and lowest for architectures with restricted tokenization (Med-Gemma). Embedding-based clustering often yields stronger separation when applied directly to the generated answers, whereas NLI-based clustering remains advantageous for LLaVA-Med and for longer, sentence-level responses. Across configurations, the VASE metric consistently provides the most robust hallucination signal, especially when paired with embedding clustering and a moderate sampling budget ($n{\approx}10$--15). Prompt design also matters: concise, label-style outputs offer clearer semantic structure than syntactically constrained one-sentence responses.

By framing hallucination detection as a geometric robustness problem shaped jointly by sampling scale, prompt structure, model architecture, and clustering strategy, HEDGE provides a principled, compute-aware foundation for evaluating multimodal reliability. The \texttt{hedge-bench} PyPI library enables reproducible and extensible benchmarking, with full code and experimental resources available at \url{https://github.com/Simula/HEDGE}.
\end{abstract}

\maketitle
\section{Introduction}

Large \glspl{vlm} have rapidly advanced multimodal reasoning, enabling free-form question answering and grounded dialogue across complex visual scenes.
Yet their expressive power comes with a critical weakness: hallucination---the generation of confident statements inconsistent with visual evidence or logical context.
Hallucination undermines the reliability of \glspl{vlm} in high-stakes domains such as clinical decision support, scientific analysis, and autonomous systems, where factual precision and interpretability are essential.

\subsection{Limitations of existing metrics}
Recent work has sought to quantify hallucination through uncertainty and consistency metrics.
\Gls{se}~\cite{SE2024} measures uncertainty over semantically grouped responses via normalized mean log-likelihoods, while RadFlag~\cite{RadFlag2024} introduces a hallucination flagger for medical \glspl{vlm} based on contradiction frequency.
\Gls{vase}~\cite{VASE2025} extends \gls{se} by contrasting entropy distributions under clean versus perturbed visual inputs, offering robustness-oriented evaluation.
Although these methods form a strong foundation, they share key limitations: (i) they rely on text-level token statistics, ignoring the geometry of the underlying multimodal representations; (ii) they often embed domain-specific heuristics that hinder cross-domain generalization; and (iii) they can be computationally fragile, particularly for pairwise \gls{nli}-based clustering.

\subsection{From token uncertainty to geometric stability}
We posit that hallucination is not merely a probabilistic irregularity but a geometric instability within the visual-linguistic manifold.
A reliable model should maintain consistent semantic neighborhoods when its visual or linguistic inputs are perturbed.
Hallucination manifests when this geometric structure fractures---when embeddings that should remain close under minor perturbations disperse in latent space.
This view reframes hallucination estimation as a problem of dense geometric entropy: measuring how the topology of the model’s representation space deforms under controlled variation.

\subsection{Overview of the framework}
We introduce \gls{hedge}, a domain-agnostic framework for hallucination detection based on representation stability.
\Gls{hedge} systematically probes a model’s multimodal manifold through dense visual perturbations and computes geometric entropy across semantically clustered responses.
It unifies two complementary clustering strategies:
(i) \gls{nli}-based clustering, which measures entailment consistency among answers~\cite{MNLI2018,DeBERTaHe2020Jun}; and
(ii) embedding-based clustering, which groups semantically aligned responses in representation space~\cite{reimers2019sentence,gao2021simcse}.
By analyzing entropy within these clusters, \gls{hedge} estimates hallucination likelihood without requiring task-specific labels or handcrafted thresholds.

\subsection{Why medical benchmarks?}
Although \gls{hedge} is not medical-specific, we evaluate it on VQA-RAD~\cite{lau2018vqarad} and KvasirVQA-x1~\cite{gautam2025kvasirvqax1} because medical VQA offers a controlled setting for isolating hallucination behavior. These datasets provide (i) short, low-ambiguity answers that reduce linguistic variance; (ii) high-resolution images with consistent visual structure, making perturbations easier to parameterize and evaluate; and (iii) established hallucination baselines---\gls{se}~\cite{SE2024}, RadFlag~\cite{RadFlag2024}, and \gls{vase}~\cite{VASE2025}---that support rigorous, reproducible comparison. We use medical imagery as a structured stress test for representation stability rather than as a domain constraint: the \gls{hedge} pipeline is domain-agnostic and can be instantiated on other VQA settings by adopting perturbations and prompting conventions suited to the target domain.

\subsection{Architectural factors in hallucination geometry}
Our analysis spans three representative \glspl{vlm} families with distinct vision encoders and fusion mechanisms:
(1) LLaVA-Med~\cite{LLaVAmed} adopts a \gls{clip} ViT-L/14 vision tower with images normalized to 336\,px and a dual-tower alignment module;
(2) Med-Gemma~\cite{MedGemma} employs a SigLIP-based encoder operating on 896$\times$896\,px crops, which are compressed into a fixed 256-token visual bottleneck; and
(3) Qwen2.5-VL~\cite{Qwen25vl} processes images at dynamically selected resolutions using a unified cross-modal transformer that preserves variable-length visual-token sequences%
\ifdefined\UseOpreSNet
\footnote{Some variants additionally support OpreSNet/OpenResNet vision towers~\cite{OpenResNet}.}%
\fi.
These architectural differences shape the local geometry of visual embeddings and thereby directly influence the detectability of hallucination under input perturbations.

\subsection{Contributions}
In this work, we contribute two main components. First, we introduce \gls{hedge}, a unified and domain-agnostic framework for hallucination detection in vision–language models that standardizes sampling, clustering, and robustness-based scoring. Second, we provide a systematic comparison of design choices within this framework, including NLI-based versus embedding-based clustering, answer-length prompting, and sampling scales. Our embedding-based clustering variants are new additions designed to compete with the NLI-based clustering used in prior work, and all reported results are produced within the same \gls{hedge} pipeline to enable controlled and reproducible comparisons.

\subsection{Takeaway}
\Gls{hedge} reframes hallucination detection from a metric-centric exercise into a study of representation stability.
By quantifying geometric entropy rather than token-level uncertainty, it exposes how architectural choices, visual tokenization density, and fusion topology govern a model’s susceptibility to hallucination---establishing a principled, scalable foundation for robust \gls{vlm} evaluation.

\section{Related Work}
\label{sec:related}

Hallucination detection connects to uncertainty modeling, semantic grouping, consistency-based evaluation, entailment reasoning, retrieval-grounded attribution, multimodal robustness, and medical reliability assessment. We briefly review these directions and, for each, indicate how they motivate the \gls{hedge} framework.

\subsection{Uncertainty and Semantic Dispersion}

Uncertainty quantification provides a core mechanism for assessing hallucination risk in both \glspl{llm} and \glspl{vlm}. Early reference-free metrics such as AvgProb, MaxProb, AvgEnt, and MaxEnt estimate token-level confidence or entropy but remain sensitive to paraphrasing and lexical variation~\cite{ReferenceFree2024}. Semantic Entropy (SE) reduces this sensitivity by modeling dispersion across semantically grouped samples~\cite{SE2024}, and Semantic Entropy Probes (SEP) approximate SE with lightweight regressors over hidden states~\cite{Kossen2024SEP}. In multimodal reasoning, VASE compares semantic entropy under clean and perturbed visual inputs~\cite{VASE2025}, and DSE introduces discretized semantic variability tailored to radiology filtering~\cite{DSE2025}. More broadly, perturbation-based UQ methods apply paraphrasing, visual occlusion, or structured noise to quantify variability~\cite{gao2024spuq, padhi2025multimodal, epistemic-uq2024}, but may suffer from calibration instability in high-dimensional visual settings~\cite{vashurin2025benchmarking}.

These works highlight the importance of semantic dispersion under perturbations, and \gls{hedge} builds on this line by computing SE, RadFlag, and VASE over semantically clustered samples within a unified perturbation and sampling protocol.

\subsection{Consistency-Based Detection}

Consistency-based detection identifies hallucinations by disagreement across sampled generations or across different models. SelfCheckGPT operationalizes self-agreement analysis over multiple generations~\cite{Manakul2023SelfCheckGPT}, while metamorphic prompting tests semantic invariance under controlled prompt modifications~\cite{MetaQA2025}. Cross-model agreement probes whether independently trained models converge on similar semantics~\cite{FinchZK2025}. In clinical applications, HalluciDoctor combines cross-consistency checks with uncertainty filtering to suppress unsafe outputs~\cite{HalluciDoctor2023}.

These approaches focus on behavioral stability at the level of outputs; \Gls{hedge} is related in spirit but instead measures stability of the underlying response distribution via geometric entropy across clustered generations.

\subsection{Entailment-Based and Logical Verification}

NLI-based methods detect hallucinations by checking whether generated content is logically supported by available evidence. FactCC~\cite{Kryscinski2020FactCC}, the contradiction-based approach of Falke et al.~\cite{Falke2019NLI}, and SummaC~\cite{Laban2022SummaC} rely on entailment models to flag unsupported statements. In biomedical and radiological settings, RadFlag~\cite{RadFlag2024} and clinical NLI systems~\cite{Romanov2018, Suhas2025ClinicalNLI} adapt this idea to domain-specific safety constraints. VASE~\cite{VASE2025} uses pairwise NLI predictions to cluster semantically equivalent outputs before computing entropy.

NLI-based clustering is interpretable but incurs quadratic comparison cost and sensitivity to paraphrasing, so \gls{hedge} retains NLI-based clustering as one option and complements it with an embedding-based alternative in the same pipeline in order to study these tradeoffs systematically.

\subsection{Semantic Grouping Approaches}

Semantic grouping underlies many hallucination metrics. Symbolic methods such as PASCAL RTE~\cite{dagan2006pascal} and entailment graphs~\cite{berant2011global} provide directional reasoning but scale poorly. Geometric embedding-based methods represent semantics in vector space using models like Word2Vec~\cite{mikolov2013efficient}, BERT~\cite{devlin2018bert}, Sentence-BERT~\cite{reimers2019sentence}, and multimodal encoders such as \gls{clip}~\cite{radford2021learning}, with visualization tools such as t-SNE~\cite{vandermaaten2008visualizing}. Hybrid strategies combine symbolic and geometric information, for example SimCSE which incorporates NLI contrastive signals~\cite{gao2021simcse}, asymmetric embedding models for directional inference~\cite{ma2018awe}, and hybrid graph–embedding frameworks that integrate structured knowledge~\cite{kapanipathi2020infusing, bastola2025hybrid, ifrim2025hybridsc}.

\Gls{hedge} leverages this literature by directly comparing a logical (NLI-based) clustering strategy with a geometric (embedding-based) strategy and by using the resulting semantic groups as the basis for its hallucination metrics.

\subsection{Retrieval-Grounded Attribution}

Retrieval-augmented methods evaluate hallucinations by aligning generated claims with retrieved evidence. Attribution surveys provide overviews of faithfulness and grounding techniques~\cite{AttributionSurvey2024}, and RAGAS defines metrics for context precision, context recall, and answer relevancy in retrieval-augmented QA~\cite{RAGAS2024}. These approaches treat external documents as the primary source of ground truth.

In contrast, \gls{hedge} targets image-grounded reasoning where the main evidence is the image itself, and therefore focuses on the internal geometric structure of model outputs under controlled visual perturbations rather than on external retrieval.

\subsection{LLM-as-a-Judge and Verifier Models}

LLM-based evaluators such as G-Eval~\cite{Liu2023GEval} use rubric-driven judgments but can exhibit bias and instability~\cite{JudgeBiasSurvey2024}. Lightweight verifier models like HalluGuard~\cite{HalluGuard2025} aim for efficiency but require supervised training and expose hallucination risk only through an additional classifier.
\Gls{hedge} uses an LLM adjudicator only to create binary hallucination labels for evaluation, and the detection method itself remains unsupervised and based on the model's own response geometry.

\subsection{Hallucination Detection in VLMs}

\subsubsection{Internal-State and Attention Probes}

Internal representations offer another view of hallucination risk. Prior work shows that fusion-layer and mid-layer activations can serve as pre-generation predictors, and that mid-layer probes can identify neurons associated with hallucination~\cite{MiddleLayerProbesCVPR2025}. Attention-based grounding diagnostics such as VADE~\cite{VADE2025} and attention calibration~\cite{AttentionCalibration2025} evaluate cross-modal alignment quality.

\Gls{hedge} is complementary to these internal probes, since it operates on generated answers and their token log-probabilities but still reflects internal structure through the embedding and clustering procedures.

\subsubsection{Vision Cross-Checks and CLIP Consistency}

External vision models such as OWL-ViT, GLIP, and SEEM enable claim verification via detector voting~\cite{DetectorVoting2025}. Caption hallucination benchmarks like CHAIR/CHAIRi~\cite{ObjectHalBench2024} and CLIP-similarity methods, including cycle consistency~\cite{CycleConsistency2025}, measure alignment between captions and visual content.

\Gls{hedge} does not rely on external detectors or CLIP scores; instead, it evaluates hallucination risk through the stability of the model's own multimodal manifold under visual perturbations.

\subsubsection{Perturbation and Counterfactual Methods}

Visual perturbations such as noise, cropping, occlusion, and blur expose grounding fragility~\cite{VASE2025}. Counterfactual VLM tests analyze semantic stability under hypothetical changes to the input~\cite{CounterfactualVLM2025}. These methods primarily focus on correctness under perturbation rather than on the structure of semantic partitions induced by the perturbations.

\Gls{hedge} explicitly treats visual perturbations as a tool to probe the geometry of the response distribution and computes geometric entropy within semantic clusters formed from clean and distorted generations.

\subsection{Medical VLM Hallucination and Uncertainty}

In medical \glspl{vlm}, hallucinations are particularly consequential due to clinical safety constraints in radiology, endoscopy, and pathology. Uncertainty-based approaches~\cite{gao2024spuq, padhi2025multimodal}, entailment-driven evaluations such as RadFlag~\cite{RadFlag2024}, and multimodal reliability frameworks all emphasize the need for robust semantic stability in high-resolution medical imaging. Medical \gls{vqa} datasets, including VQA-RAD~\cite{lau2018vqarad}, MIMIC-Diff-VQA~\cite{MIMICDiffVQA2023}, HALT-MedVQA~\cite{HALTMedVQA2024}, Kvasir-VQA~\cite{gautam2024kvasirvqa}, and Kvasir-VQA-x1~\cite{gautam2025kvasirvqax1}, support domain-specific evaluation of grounding and uncertainty.

\Gls{hedge} instantiates its framework on VQA-RAD and KvasirVQA-x1 to obtain controlled medical testbeds for representation stability, while remaining domain-agnostic in its design.

\subsection{Benchmarks for Hallucination Detection}

Text hallucination datasets such as HaluEval~\cite{HaluEval2022}, HaluEval 2.0~\cite{HaluEval2024}, HaDes~\cite{HaDes2023}, FACTOID~\cite{FACTOID2025}, HalluLens~\cite{Bang2025HalluLens}, HalluEntity~\cite{HalluEntity2024}, FACTCHD~\cite{FACTCHD2024}, MedHal~\cite{MedHal2023}, and FreshQA~\cite{FreshQA2024} focus on short-form factual errors. Long-form factuality datasets, including WildHallucinations~\cite{WildHallucinations2024}, LongFact~\cite{LongFact2024}, PopQA~\cite{PopQA2023}, and SimpleQA~\cite{SimpleQA2022}, target sustained factual consistency. For summarization, the SummaC Suite~\cite{Laban2022SummaC} and FRANK~\cite{FRANK2021} capture summary hallucinations. Dialogue datasets such as FaithDial~\cite{FaithDial2022}, DialFact~\cite{DialFact2022}, DiaHalu~\cite{DiaHalu2024}, HalluDial~\cite{HalluDial2024}, and TRUE/ScreenEval subsets~\cite{TRUE2021} cover conversational hallucinations.

VLM hallucination benchmarks, including POPE~\cite{POPE2023}, H-POPE~\cite{HPOPE2024}, THRONE~\cite{THRONE2024}, HallusionBench~\cite{HallusionBench2023}, AMBER~\cite{AMBER2024}, ODE~\cite{ODE2024}, ALOHa~\cite{ALOHa2024}, OpenCHAIR~\cite{OpenCHAIR2024}, MOCHa~\cite{MOCHa2024}, Polaris/Polos~\cite{Polos2024}, CapArena~\cite{CapArena2024}, MHaluBench~\cite{MHaluBench2024}, and Object-HalBench~\cite{ObjectHalBench2024}, examine object hallucination, grounding, and factuality across modalities, and survey papers summarize taxonomies and evaluation challenges~\cite{ACMSurvey2025, MMSurvey2025, AHESurvey2024, RAGSurvey2025}.

\Gls{hedge} is designed to sit on top of such benchmarks by providing a standardized pipeline for sampling, clustering, and robustness-based scoring that can be applied without task-specific labels.

\subsection{Positioning HEDGE}

Across these threads, hallucination detection is usually framed as uncertainty estimation~\cite{SE2024, Kossen2024SEP, VASE2025, DSE2025}, agreement analysis~\cite{Manakul2023SelfCheckGPT, MetaQA2025, FinchZK2025}, entailment reasoning~\cite{Kryscinski2020FactCC, Laban2022SummaC, RadFlag2024}, attention or grounding diagnostics~\cite{VADE2025, AttentionCalibration2025}, or external alignment with detectors and retrieval~\cite{ObjectHalBench2024, radford2021learning}. What is less explored is how visual perturbation, sampling scale, semantic grouping strategy, tokenization density, and fusion topology jointly shape the geometry of a model's response distribution.

\Gls{hedge} addresses this gap by defining hallucination detection as stability of visual–linguistic geometric structure under controlled perturbations, sampling balanced clean and noisy generations, applying both NLI-based and embedding-based clustering, and computing SE, RadFlag, and VASE over shared semantic partitions so that architectural and data factors can be disentangled from metric choice.

\section{Methodology}
\label{sec:methodology}

\begin{figure*}[htb]
    \includegraphics[width=1\textwidth]{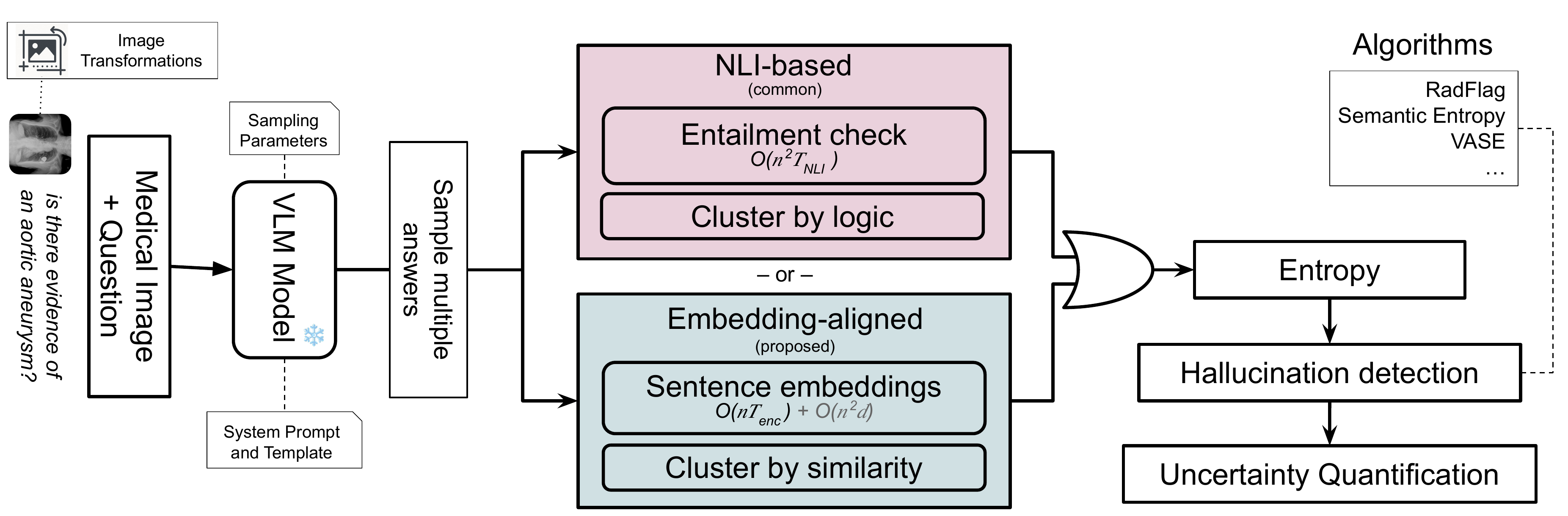}
    \centering
    \caption{Overview of the proposed \gls{hedge} framework for hallucination detection in visual question answering (VQA).
    A vision--language model generates multiple answers per image--question pair, which are grouped via two strategies:
    natural language inference (NLI)-based logical clustering and embedding-aligned semantic clustering.
    Entropy within these groups quantifies uncertainty, enabling hallucination detection through metrics such as
    RadFlag, Semantic Entropy, and VASE.}
    \label{fig:teaser}
\end{figure*}

\Gls{hedge} is a modular
hallucination benchmarking framework, not a single metric. It unifies answer
sampling, semantic clustering, threshold optimization, and hallucination
scoring within a single, architecture-agnostic pipeline
(Figure~\ref{fig:teaser}). The framework standardizes (i) answer sampling under
different prompt and temperature settings, (ii) semantic clustering using either
NLI-based or embedding-based backends, and (iii) scoring with established
hallucination metrics such as RadFlag, SE, and VASE, along with our
embedding-aligned variants. By formulating hallucination detection as a
geometric robustness problem, \gls{hedge} enables consistent and comparable
evaluation across multimodal models. An open-source implementation,
\href{https://pypi.org/project/hedge-bench}{\texttt{hedge-bench}}, provides all
components needed to reproduce and extend our experiments.

\subsection{Overview of the Pipeline}
\label{subsec:pipeline_overview}

\subsubsection{Answer Generation}
For each image, we generate one low-temperature baseline answer $A_0$ and multiple high-temperature responses under two visual conditions: clean and noisy~\cite{TemperatureamplingRenze2024}.
Clean responses ($A_1$, $A_2$, $\ldots$, $A_n$) are generated from the original image, while noisy responses ($N_1$, $N_2$, $\ldots$, $N_n$) are produced from visually distorted variants.
The number of distortions ($n$) can vary depending on the experimental setup.
Mean token log-probabilities are recorded for both conditions, denoted $\log \ell_{\text{clean}}$ and $\log \ell_{\text{noisy}}$.

\subsubsection{Sequence Assembly}
All responses are arranged as $\texttt{seq\_input} = [A_0] + [A_1..A_n] + [N_1..N_n]$.
In the Answer+Question mode, the question text is prepended to each answer only during clustering, while log-prob arrays remain aligned with their respective responses.

\subsubsection{Semantic Clustering}
Clustering is applied to $\texttt{seq\_input}$ with index mapping:
$[0]\!\rightarrow\!A_0$, $[1{:}n{+}1]\!\rightarrow\!A_1..A_n$ (clean), and $[n{+}1{:}2n{+}1]\!\rightarrow\!N_1..N_n$ (noisy).  
Each response $A_i$ (or $N_i$) is assigned a discrete cluster identifier: $c_i = \text{cluster}(A_i)$.
Two clustering strategies are compared (Sec.~\ref{subsec:nli_vs_embedding}).

\subsubsection{Metric Computation}
Given mean token log-likelihoods $\log \ell_i$ and their corresponding semantic cluster identifiers $c_i$, 
we first compute the semantic distribution and semantic entropy. 
The semantic distribution over clusters is defined as:
\begin{equation}
s_j = 
\frac{
    \exp\!\left(\sum_{i: c_i = j} \exp(\log \ell_i - \max_k \log \ell_k)\right)
}{
    \sum_{m} \exp\!\left(\sum_{i: c_i = m} \exp(\log \ell_i - \max_k \log \ell_k)\right)
},
\label{eq:semantic-distribution}
\end{equation}
and the corresponding semantic entropy is:
\begin{equation}
\mathrm{SE}(\mathbf{s}) = -\sum_{j} s_j \log s_j~\cite{Shannon1948}.
\label{eq:semantic-entropy}
\end{equation}

We denote the semantic distributions from clean and noisy generations as 
$\mathbf{s}_{\text{clean}}$ and $\mathbf{s}_{\text{noisy}}$, respectively. 
Based on these, we define three hallucination detection metrics.

\Gls{se} quantifies uncertainty over semantic clusters derived from clean generations~\cite{SE2024}:
\begin{equation}
\mathrm{SE} = \mathrm{SE}(\mathbf{s}_{\text{clean}}).
\label{eq:se}
\end{equation}

RadFlag measures the proportion of high-temperature clean generations whose semantic cluster differs from that of the baseline, indicating semantic inconsistency or hallucination~\cite{RadFlag2024}:
\begin{equation}
\mathrm{RadFlag} = 1 - \frac{1}{n}\sum_{i=1}^{n}\mathbf{1}[c_i = c_0].
\label{eq:radflag}
\end{equation}

\Gls{vase} extends semantic entropy by capturing the stability gap between clean and noisy conditions~\cite{VASE2025}:
\begin{equation}
\mathrm{VASE} = 
\mathrm{SE}\!\left(
\operatorname{softmax}\!\left(
\mathbf{s}_{\text{clean}} + 
\alpha(\mathbf{s}_{\text{clean}} - \mathbf{s}_{\text{noisy}})
\right)
\right),
\label{eq:vase}
\end{equation}
where $\alpha$ is a scaling factor controlling the influence of the clean--noisy semantic gap 
(we set $\alpha = 1$ in all experiments, following ~\cite{VASE2025}).

Increasing the number of distortions enlarges the noisy sample pool, while the clean high-temperature pool is proportionally expanded to maintain balance, thereby affecting all three metrics:
$\mathrm{SE}$ (Equation~\ref{eq:se}), $\mathrm{RadFlag}$ (Equation~\ref{eq:radflag}), and $\mathrm{VASE}$ (Equation~\ref{eq:vase}).
Thus, any experiment that varies $n$ jointly modulates sampling scale and visual perturbation strength, and the downstream analyses reflect this coupled effect.

\subsection{Semantic Clustering: NLI vs. Embedding}
\label{subsec:nli_vs_embedding}

To capture semantically coherent response clusters, we explore two complementary strategies: one grounded in logical entailment reasoning (\gls{nli}-based) and another in geometric similarity (embedding-based).

\textbf{(a) NLI-Based Clustering}
Given responses $S=\{t_1,\dots,t_m\}$, a \gls{mnli} model~\cite{MNLI2018} predicts $y_{ij}\in\{\text{entails},\text{contradicts},\text{neutral}\}$ for each pair $(t_i,t_j)$.
A directed entailment graph $G_{\text{ent}}$ is built with edges $E_{\text{ent}}=\{(i,j)|y_{ij}=\text{entails}\}$.  
Clusters are derived via mutual entailment closure:
$
i \sim j \iff (i,j)\in E_{\text{ent}} \wedge (j,i)\in E_{\text{ent}},
$
followed by union–find transitive merging.  
Contradiction filtering prevents antonym merging.  

\textbf{(b) Embedding-Based Clustering}
Each response is embedded via a sentence encoder~\cite{reimers2019sentence}, normalized to unit length, with pairwise cosine similarity $\sigma_{ij}=\text{x}_i^\top\text{x}_j$.  
An undirected graph $G_{\text{sim}}$ is built via:
$
\sigma_{ij}\ge\tau \quad \text{or} \quad j\in\text{kNN}(i),
$
and clusters correspond to connected components.  
The similarity threshold $\tau$ controls granularity and is automatically tuned to maximize \gls{rocauc}~\cite{fawcett2006ROC}.
This method scales as $\mathcal{O}(n T_{\text{enc}}) + \mathcal{O}(n^{2} d)$; 
although quadratic in $n$, the $\mathcal{O}(n^{2} d)$ component consists only of 
inexpensive vector dot products, making the approach robust for large candidate sets. 
Here, $n$ denotes the number of responses, $d$ is the embedding dimension, and
$T_{\text{enc}}$ is the cost of a single embedding forward pass. For comparison, the NLI-based alternative requires $\mathcal{O}(n^{2} T_{\text{NLI}})$
pairwise model inferences, where $T_{\text{NLI}}$ is a full transformer forward pass,
making it substantially more expensive at scale.
A concise comparison of both clustering strategies is provided in Table~\ref{tab:nli_vs_embedding}.

\begin{table}[h!]
\footnotesize
\centering
\caption{Comparison of \gls{nli}- and embedding-based semantic clustering. 
$n$ = number of responses; $d$ = embedding dimension; 
$T_{\text{NLI}}$ and $T_{\text{enc}}$ are single forward-pass costs.}
\label{tab:nli_vs_embedding}
\setlength{\tabcolsep}{4pt}
\renewcommand{\arraystretch}{1.12}
\begin{tabular}{p{2cm}p{3cm}p{3cm}}
\toprule
Property & NLI-Based & Embedding-Based \\
\midrule
Semantic basis & Logical entailment & Geometric similarity \\
Relation type & Directed, asymmetric & Undirected, symmetric \\
Contradiction & Explicitly modeled & Not explicit \\
Transitivity & Mutually enforced & Emergent \\

Complexity
& $\mathcal{O}(n^{2} T_{\text{NLI}})$ 
& $\mathcal{O}(n T_{\text{enc}}) {\color{gray} + \mathcal{O}(n^{2} d)}$ \\

Scalability & Low & High \\
Interpretability & High & Moderate \\
\bottomrule
\end{tabular}
\end{table}

\subsection{Hallucination Metrics}

\Gls{se}~\cite{SE2024} quantifies uncertainty across semantically grouped answers using normalized mean log-likelihoods.  
RadFlag~\cite{RadFlag2024} measures the proportion of contradictory or inconsistent responses within a cluster.  
\Gls{vase}~\cite{VASE2025} extends \gls{se} by contrasting clean and noisy \gls{se} distributions, thus evaluating robustness to visual perturbations.

All metrics are computed in a batched manner with stable mean-log-prob aggregation for scalability.

\subsection{Answer-Length Prompt Configurations}
\label{subsec:answer_length}
\begin{figure}[h]
\centering
\lstset{
    basicstyle=\ttfamily\scriptsize,
    breaklines=true,
    breakatwhitespace=false,
    columns=fullflexible,
    keepspaces=true,
    showstringspaces=false,
    frame=none,
}
\begin{lstlisting}[caption={Prompt templates for each answer-length prompt configuration.}, label={lst:answer_length_prompts}]
{
"default": [
    { "role": "system",
    "content": "You are a medical image analysis expert."
    },{
    "role": "user",
    "content": "<image> Answer this question as concisely as possible based on the provided image: {question}"
    }],
"minimal-label": [
    {"role": "system",
    "content": "You are a medical vision-language assistant. Given a medical image and a clinical question, provide only the minimal, clinically correct answer. Answers may be 'yes', 'no', or a short medically relevant label (e.g., modality, anatomy, finding). Do not generate full sentences, explanations, or extra text - only output exactly the expected label."
    },
    {"role": "user",
    "content": "<image> Question: {question}"
    }],
"one-sentence": [
    {"role": "system",
    "content": "You are a medical image analysis expert; always respond in no more than a single sentence."
    },
    {"role": "user",
    "content": "<image> Answer this question as concisely as possible based on the provided image: {question}"
    }],
"clinical-phrase": [
    {"role": "system",
    "content": "You are a medical image analysis expert. Given an image and a question, provide a concise and accurate response. The answer should be a short phrase, slightly longer than a single label if needed, but not a full grammatical sentence and without a period at the end. Avoid explanations or extra text - only give the short direct answer."
    },
    {"role": "user",
    "content": "<image> Question: {question}"
    }]
}
\end{lstlisting}
\end{figure}

To probe the effect of linguistic verbosity on hallucination behavior, we define four prompt configurations—default, one-sentence, clinical-phrase, and minimal-label.
Each prompt configuration modifies the granularity of model outputs while preserving the same underlying image–question pair.  
The exact system–user prompt templates used to elicit responses under each prompt configuration are shown in Listing~\ref{lst:answer_length_prompts}.  
These templates progressively constrain the allowable response length, ranging from natural free-form answers (default) to minimal diagnostic labels (minimal-label), allowing us to analyze how linguistic compression affects hallucination tendencies across models and datasets~\cite{VASE2025}.

\section{Experiments}
\label{sec:experiments}

This section details how we instantiate the \gls{hedge} framework for our experiments, including datasets, models, evaluation protocol, and experimental setup for hallucination detection across both VQA-RAD and KvasirVQA-x1.
All experiments are conducted within the unified pipeline described in Section~\ref{sec:methodology}, combining answer generation, semantic clustering, threshold optimization, and metric computation.

\subsection{Datasets}
\label{subsec:datasets}

The VQA-RAD dataset~\cite{lau2018vqarad} contains radiology images paired with 451 free-text clinical questions spanning modalities, anatomical regions, and diagnostic findings.
We use the official test split for evaluation and retain the original question distribution.
The KvasirVQA-x1 dataset~\cite{gautam2025kvasirvqax1} focuses on \gls{vqa} in gastrointestinal endoscopy and includes multiple complexity levels and visual variability settings.
For our experiments, we filter and use a subset of 500 representative test samples at complexity level~1, without additional perturbations or augmentations.

\subsection{Models}
\label{subsec:models}

We evaluate three representative vision–language models under identical inference configurations: LLaVA-Med-v1.5-Mistral-7B~\cite{LLaVAmed}, which is instruction-tuned on medical and visual QA data; Med-Gemma-4B-IT~\cite{MedGemma}, Google’s multimodal Gemma variant with medical instruction tuning; and Qwen2.5-VL-7B-Instruct~\cite{Qwen25vl}, a strong general-purpose multimodal model with instruction-following capabilities.

Images are preprocessed according to each model’s native vision encoder input.
LLaVA-Med keeps aspect ratio then center-crops to 336×336, MedGemma stretches every image to a fixed 896×896 square, and Qwen2.5-VL preserves the full image and aspect ratio.
All images are normalized to each model’s pretrained statistics.
All models are evaluated zero-shot without any fine-tuning.
We follow each model’s official preprocessing, tokenizer, and image encoder configurations.
All experiments are run on a single NVIDIA A100 GPU (80~GB) in BF16 precision.

\subsection{Answer-Length Configurations}
\label{subsec:answer_regimes}

Each question–image pair is evaluated under four prompting templates:
default, one-sentence, clinical-phrase, and minimal-label.
The same image and question are reused while only the prompt style is changed.
All prompt configurations employ sampling with temperature values \(T = 0.1\) (low-temperature baseline) and \(T = 1.0\) (high-temperature sampling), using each model's default decoding parameters.

\subsection{Answer Generation Protocol}
\label{subsec:answer_generation}

For each question, one low-temperature baseline ($A_0$) and multiple high-temperature responses are generated under clean and noisy visual conditions.
The joint sampling scale per image ($n$ clean high-temperature answers and $n$ perturbed counterparts) is varied from 1 to 30 (\texttt{[1-10, 15, 20, 25, 30]}).
Distorted variants include random affine transforms (rotation $\pm10^\circ$, translation $\pm10\%$, scale $\pm10\%$), color jitter (brightness/contrast $\pm20\%$, saturation $\pm5\%$, hue $\pm0.02$), and additive Gaussian ($\sigma=0.07$) and Poisson (scale $=0.014$) noise implemented via the Albumentations library~\cite{Buslaev2018Sep}.
Each generated answer includes its mean token log-probability, later used for \gls{se} and \gls{vase} computation.

\subsection{Ground-Truth Hallucination Labels}
\label{subsec:hallucination_labels}

Ground-truth hallucination supervision is generated using a large language model acting as a strict medical adjudicator.
For each item, the adjudicator receives the question, the clinically verified correct answer, and the model’s generated answer.
In the KvasirVQA-x1 test set, the expected answer is provided as part of the question description, offering additional clinical context for evaluation.
The adjudicator assesses factual and semantic consistency between the reference and generated answers, focusing strictly on clinical correctness.
A response is marked as hallucinated when the generated answer introduces false, contradictory, or medically inaccurate information; minor wording differences are tolerated if the underlying meaning is preserved.
Each evaluation produces a single binary decision (\texttt{supported} vs.\ \texttt{hallucinated}) and may include a brief rationale.
We employ Qwen3-30B-A3B~\cite{Qwen3Yang2025May} as the adjudicator model, with few-shot calibration to enhance evaluation reliability.

\subsection{Metric Evaluation and Comparison}
\label{subsec:metric_evaluation}

We evaluate three hallucination metrics: RadFlag~\cite{RadFlag2024}, which measures contradiction frequency among clustered responses; \gls{se}~\cite{SE2024}, which captures the entropy of mean log-probabilities within semantic clusters; and \gls{vase}~\cite{VASE2025}, which quantifies the robustness gap between clean and noisy \gls{se} distributions.
Each metric is computed under both \gls{nli}-based and embedding-based clustering to keep comparisons consistent across sampling regimes.
Embedding-based clustering uses SentenceTransformer embeddings~\cite{reimers2019sentence} with threshold $\tau$ optimized by Optuna~\cite{Akiba2019Jul} to maximize \gls{rocauc}, whereas \gls{nli}-based clustering relies on a DeBERTa model (\href{https://huggingface.co/microsoft/deberta-v2-xlarge-mnli}{deberta-v2-xlarge-mnli}) fine-tuned on \gls{mnli} entailment predictions~\cite{DeBERTaHe2020Jun} with contradiction filtering and transitive merging.
Implementation uses batched inference ($B=512$), deduplication, and GPU-safe cleanup to manage the $\mathcal{O}(n^2T_{\text{NLI}})$ complexity, while all metrics are computed in batches with stable mean-log-prob aggregation.
We report \gls{rocauc} against the binary hallucination labels as the main quantitative measure~\cite{fabian2011scikit}.

\subsection{Experimental Protocol}
\label{subsec:protocol}

Each question is evaluated under all four prompt configurations and both input configurations (Answer-only and Answer+Question).
We aggregate results across the entire test split without reweighting.
All hyperparameters remain fixed unless explicitly swept in ablation studies.

\subsubsection{Threshold Optimization}
Optuna-based hyperparameter tuning~\cite{Akiba2019Jul} is performed to find the optimal embedding similarity threshold $\tau^*$ that maximizes \gls{rocauc}.
Each optimization run executes full clustering and metric recomputation, with search space $\tau \in [0.8, 0.99]$ and $N = 20$ trials.

\subsubsection{Reproducibility}
All experiments use controlled batching and cached embeddings/\gls{nli} predictions.
Intermediate artifacts (answers, embeddings, logits, and adjudicator labels) are shared for full reproducibility.

\subsubsection{Runtime and Complexity}
\Gls{nli} clustering scales as $\mathcal{O}(n^{2} T_{\text{NLI}})$ pairwise model inferences, 
while embedding-based clustering scales as 
$\mathcal{O}(n T_{\text{enc}}) + \mathcal{O}(n^{2} d)$, where the quadratic term consists 
only of inexpensive dot products.
Typical runtime per dataset involves $K=500$ (KvasirVQA-x1) or $K=451$ (VQA-RAD test) samples and $n \in [1,30]$ paired clean-and-perturbed generations per image.

\subsection{Baselines and Ablations}
\label{subsec:baselines}

Within \gls{hedge}, we closely follow the official \gls{vase} and RadFlag implementations to ensure consistency with prior work, matching their sampling settings, perturbation levels, and aggregation protocols as closely as possible. This allows our embedding-aligned variants and NLI/embedding clustering choices to be compared to these baselines under a single unified pipeline.
Our analysis juxtaposes the official RadFlag, \gls{se}, and \gls{vase} scores with embedding-aligned variants (\gls{se}$_\text{emb}$, RadFlag$_\text{emb}$, and \gls{vase}$_\text{emb}$ tuned via Optuna) while contrasting \gls{nli}-driven and embedding-driven clustering across models and prompt regimes.
Ablations vary the sampling scale and perturbation budget, the inclusion of question text during clustering, the prompt configuration length, and the embedding similarity threshold $\tau$ to reveal how each factor influences performance.
Results are reported as mean \glspl{rocauc} across models, datasets, and prompt configurations, with each experiment logging metric outputs, the selected $\tau^*$, per-model \glspl{auc}, and per-sampling-scale trends for reproducibility.

\begin{table*}[h!]
\centering
\footnotesize
\caption{
ROC-AUC score comparison of hallucination detection methods across answer-length prompt configurations for the VQA-RAD and KvasirVQA-x1 test sets. Bold values within each row indicate the better performer for that model. Two input configurations are evaluated: A+Q (\textit{Answer + Question passed}) — where the question is appended to the model-generated answer before being processed by the NLI or sentence embedding method, and AO (\textit{Answer Only passed}) — where only the answer text is used. Results are reported for LLaVA-Med, Med-Gemma, and Qwen2.5-VL. All experiments are conducted using 10 visual distortion variants per question.
}
\label{tab:auc_comparison_full_split}
\setlength{\tabcolsep}{5pt}
\renewcommand{\arraystretch}{1.15}
\begin{tabular}{llcccccccccccc}
\toprule
\multirow{2}{*}{\textbf{Answer Length }} &
\multirow{2}{*}{\textbf{Method}} &
\multicolumn{6}{c}{\textbf{A+Q (Answer + Question passed)}} &
\multicolumn{6}{c}{\textbf{AO (Answer Only passed)}} \\
\cmidrule(lr){3-8} \cmidrule(lr){9-14}
& &
\multicolumn{2}{c}{\textbf{LLaVA-Med}} &
\multicolumn{2}{c}{\textbf{Med-Gemma}} &
\multicolumn{2}{c}{\textbf{Qwen2.5-VL}} &
\multicolumn{2}{c}{\textbf{LLaVA-Med}} &
\multicolumn{2}{c}{\textbf{Med-Gemma}} &
\multicolumn{2}{c}{\textbf{Qwen2.5-VL}} \\
\cmidrule(lr){3-4} \cmidrule(lr){5-6} \cmidrule(lr){7-8}
\cmidrule(lr){9-10} \cmidrule(lr){11-12} \cmidrule(lr){13-14}
& & \textbf{NLI} & \textbf{Embed.} & \textbf{NLI} & \textbf{Embed.} & \textbf{NLI} & \textbf{Embed.} & \textbf{NLI} & \textbf{Embed.} & \textbf{NLI} & \textbf{Embed.} & \textbf{NLI} & \textbf{Embed.} \\
\midrule

\multicolumn{14}{l}{\textbf{VQA-RAD Test Set}} \\
\midrule
\textbf{minimal-label} &
SE & 0.734 & 0.701 & 0.673 & 0.681 & 0.772 & 0.651 & 0.739 & 0.702 & 0.677 & 0.721 & 0.764 & \textbf{0.778} \\
 &
RadFlag & 0.726 & 0.704 & 0.665 & 0.680 & 0.757 & 0.654 & 0.725 & 0.703 & 0.671 & 0.716 & 0.747 & \textbf{0.767} \\
 &
VASE & 0.725 & 0.712 & 0.702 & 0.726 & 0.750 & 0.666 & 0.734 & 0.707 & 0.717 & \textbf{0.757} & 0.750 & 0.746 \\

\textbf{clinical-phrase} &
SE & 0.728 & 0.693 & 0.700 & 0.718 & \textbf{0.776} & 0.685 & 0.731 & 0.682 & 0.650 & 0.663 & 0.738 & 0.724 \\
 &
RadFlag & 0.738 & 0.698 & 0.705 & 0.712 & \textbf{0.778} & 0.675 & 0.737 & 0.687 & 0.651 & 0.647 & 0.745 & 0.730 \\
 &
VASE & 0.720 & 0.704 & 0.762 & 0.750 & \textbf{0.767} & 0.683 & 0.721 & 0.691 & 0.725 & 0.707 & 0.739 & 0.721 \\

\textbf{default} &
SE & 0.753 & 0.716 & 0.690 & 0.666 & \textbf{0.755} & 0.686 & 0.751 & 0.698 & 0.689 & 0.652 & 0.733 & 0.721 \\
 &
RadFlag & 0.748 & 0.719 & 0.670 & 0.651 & \textbf{0.756} & 0.683 & 0.746 & 0.684 & 0.667 & 0.623 & 0.740 & 0.713 \\
 &
VASE & 0.725 & 0.709 & 0.714 & 0.678 & \textbf{0.756} & 0.707 & 0.732 & 0.690 & 0.712 & 0.669 & 0.733 & 0.731 \\

\textbf{one-sentence} &
SE & 0.740 & 0.726 & 0.675 & 0.649 & 0.706 & 0.631 & \textbf{0.752} & 0.722 & 0.657 & 0.612 & 0.707 & 0.663 \\
 &
RadFlag & 0.742 & 0.698 & 0.675 & 0.634 & 0.695 & 0.644 & \textbf{0.744} & 0.698 & 0.655 & 0.608 & 0.697 & 0.644 \\
 &
VASE & 0.731 & 0.723 & 0.705 & 0.678 & 0.728 & 0.658 & \textbf{0.736} & 0.717 & 0.696 & 0.644 & 0.730 & 0.681 \\

\midrule
\multicolumn{14}{l}{\textbf{KvasirVQA-x1 Test Set}} \\
\midrule
\textbf{minimal-label} &
SE & 0.746 & 0.706 & 0.522 & 0.643 & 0.742 & \textbf{0.781} & 0.761 & 0.650 & 0.555 & 0.592 & 0.741 & 0.768 \\
 &
RadFlag & 0.741 & 0.715 & 0.524 & 0.642 & 0.746 & \textbf{0.787} & 0.750 & 0.656 & 0.556 & 0.589 & 0.745 & 0.776 \\
 &
VASE & 0.732 & 0.705 & 0.531 & 0.685 & 0.783 & \textbf{0.809} & 0.744 & 0.652 & 0.562 & 0.624 & 0.783 & 0.800 \\

\textbf{clinical-phrase} &
SE & 0.783 & 0.705 & 0.695 & 0.707 & \textbf{0.848} & 0.847 & 0.797 & 0.654 & 0.634 & 0.586 & 0.814 & 0.823 \\
 &
RadFlag & 0.766 & 0.716 & 0.697 & 0.694 & 0.839 & \textbf{0.844} & 0.784 & 0.652 & 0.618 & 0.585 & 0.802 & 0.818 \\
 &
VASE & 0.779 & 0.711 & 0.717 & 0.738 & 0.876 & \textbf{0.876} & 0.786 & 0.659 & 0.669 & 0.622 & 0.845 & 0.868 \\

\textbf{default} &
SE & 0.778 & 0.698 & 0.610 & 0.629 & 0.875 & 0.868 & 0.786 & 0.665 & 0.629 & 0.602 & 0.866 & \textbf{0.876} \\
 &
RadFlag & 0.768 & 0.720 & 0.606 & 0.589 & 0.846 & 0.837 & 0.782 & 0.686 & 0.625 & 0.578 & 0.839 & \textbf{0.853} \\
 &
VASE & 0.779 & 0.704 & 0.627 & 0.633 & 0.879 & 0.878 & 0.782 & 0.667 & 0.638 & 0.611 & 0.877 & \textbf{0.891} \\

\textbf{one-sentence} &
SE & 0.768 & 0.689 & 0.659 & 0.653 & \textbf{0.843} & 0.794 & 0.784 & 0.648 & 0.670 & 0.594 & 0.833 & 0.806 \\
 &
RadFlag & 0.756 & 0.709 & 0.649 & 0.632 & 0.812 & 0.787 & 0.774 & 0.647 & 0.660 & 0.569 & \textbf{0.818} & 0.794 \\
 &
VASE & 0.770 & 0.698 & 0.679 & 0.672 & \textbf{0.851} & 0.825 & 0.781 & 0.649 & 0.688 & 0.617 & 0.841 & 0.819 \\

\bottomrule
\end{tabular}

\end{table*}

\section{Results}
\label{sec:results}
We evaluate three hallucination metrics—RadFlag, \gls{se}, and \gls{vase}—across two datasets (VQA-RAD, KvasirVQA-x1), three VLMs (LLaVA-Med, Med-Gemma, Qwen2.5-VL), two clustering strategies (\gls{nli}- vs.\ embedding-based), two input modes (Answer+Question vs.\ Answer Only), and four answer-length prompt configurations (default, clinical-phrase, one-sentence, minimal-label).
Table~\ref{tab:auc_comparison_full_split} summarizes ROC-AUC scores for all models and datasets; Table~\ref{tab:sample_variations} and Figure~\ref{fig:auc_vs_distortion} analyze the effect of sampling scale $n$ for Qwen2.5-VL-7B-Instruct on VQA-RAD.

We treat differences of $0.01$–$0.02$ ROC-AUC as small and potentially within sampling noise; our conclusions are based on robust patterns across rows and blocks rather than single cells.

\subsection{Model Ranking and Dataset Difficulty}
\label{subsec:results_models_datasets}

Across all metrics and configurations in Table~\ref{tab:auc_comparison_full_split}, we observe a consistent model ranking: 

$\text{Qwen2.5-VL} \;>\; \text{LLaVA-Med} \;>\; \text{Med-Gemma}.$
Qwen2.5-VL often reaches $0.85$–$0.89$ ROC-AUC on KvasirVQA-x1 (e.g., default / VASE), LLaVA-Med falls in the $\sim0.70$–$0.78$ range, and Med-Gemma frequently drops toward $0.52$–$0.60$ on KvasirVQA-x1.

Hallucination detection is consistently easier on KvasirVQA-x1 than VQA-RAD.
For the same model and metric, AUCs on KvasirVQA-x1 are typically $0.1$–$0.2$ higher than on VQA-RAD.

\subsection{Effect of Answer-Length Prompts}
\label{subsec:results_answer_length}

Prompting strongly influences detectability.
\textit{Default and clinical-phrase} prompts are the most reliable overall, often yielding the highest AUCs across models and datasets.
\textit{Minimal-label} prompts work well for strong models (e.g., Qwen2.5-VL reaches $\sim0.80$–$0.81$ VASE on KvasirVQA-x1), but performance degrades more sharply for Med-Gemma.
\textit{One-sentence} prompts are the hardest regime: they systematically underperform default and clinical-phrase, especially on VQA-RAD.

In short, \Gls{hedge} works best when answers are concise but semantically expressive.
Overly compressed (minimal-label for weaker models) or strictly templated (one-sentence) prompts reduce the semantic diversity that the metrics exploit.

\subsection{Clustering Strategy: NLI vs.\ Embedding}
\label{subsec:results_clustering}

Neither clustering strategy dominates globally.

\subsubsection{VQA-RAD}
For Qwen2.5-VL and LLaVA-Med, \gls{nli}-based clustering often wins under longer prompts (default, clinical-phrase), particularly in the Answer+Question mode (e.g., Qwen2.5-VL, default, A+Q: NLI VASE $0.756$ vs.\ embedding $0.707$).

\subsubsection{KvasirVQA-x1}
For Qwen2.5-VL, embedding-based clustering is competitive or superior under minimal-label and many clinical-phrase / default AO settings (e.g., minimal-label AO VASE: $0.783$ NLI vs.\ $0.800$ embedding).
For the one-sentence prompt, NLI clearly dominates for Qwen2.5-VL (e.g., A+Q VASE: $0.851$ NLI vs.\ $0.825$ embedding, AO $0.841$ vs.\ $0.819$).

For LLaVA-Med on KvasirVQA-x1, NLI wins almost uniformly (e.g., minimal-label A+Q VASE: $0.732$ NLI vs.\ $0.705$ embedding across metrics and prompts).
Med-Gemma shows a genuinely mixed pattern: some rows favour NLI, others favour embeddings.

Embedding-based clustering uses a tuned similarity threshold $\tau$ (selected on a validation split and fixed for test); NLI uses a fixed MNLI-style model.
All NLI–embedding comparisons should be read with this calibration asymmetry in mind.

\subsection{Answer-Only vs.\ Answer+Question}
\label{subsec:results_input_modes}

The impact of concatenating the question follows a clear pattern.
Minimal-label answers benefit from Answer+Question because the question disambiguates otherwise identical short labels (e.g., “yes”, “polyp”), and the effect is most pronounced for NLI-based clustering.
Verbose answers favour Answer Only; for default and one-sentence prompts the appended question acts as a long shared prefix that suppresses contrast in embedding space, particularly for embedding-based clustering.
Thus, use A+Q when answers are semantically underspecified and stick with AO when the answer text already provides sufficient context.

\subsection{Metric Comparison: VASE vs.\ SE vs.\ RadFlag}
\label{subsec:results_metrics}

Across both datasets, three patterns stand out.
First, VASE is usually strongest: in many of the best configurations—especially for Qwen2.5-VL on KvasirVQA-x1—it achieves the highest AUC (e.g., default AO: VASE $0.891$ vs.\ SE $0.876$, RadFlag $0.853$).
Second, SE remains competitive and can occasionally take the lead; on several VQA-RAD configurations (e.g., LLaVA-Med, default, A+Q), SE slightly exceeds VASE (0.753 vs.\ 0.725), indicating that clean-only entropy can match robust metrics when clean–noisy separation is weak.
Third, RadFlag tends to trail SE and VASE, especially when hallucinations manifest as subtle semantic drifts rather than hard contradictions, yet it remains attractive at very small sampling budgets because of its low computational cost.
There is no absolute hierarchy, but “VASE $\ge$ SE $\ge$ RadFlag” serves as a reliable heuristic in the strongest regimes.

\subsection{Sampling Scale: How Many Generations Are Needed?}
\label{subsec:results_sampling_scale}

\begin{figure*}[h!]
    \centering
\includegraphics[width=\textwidth]{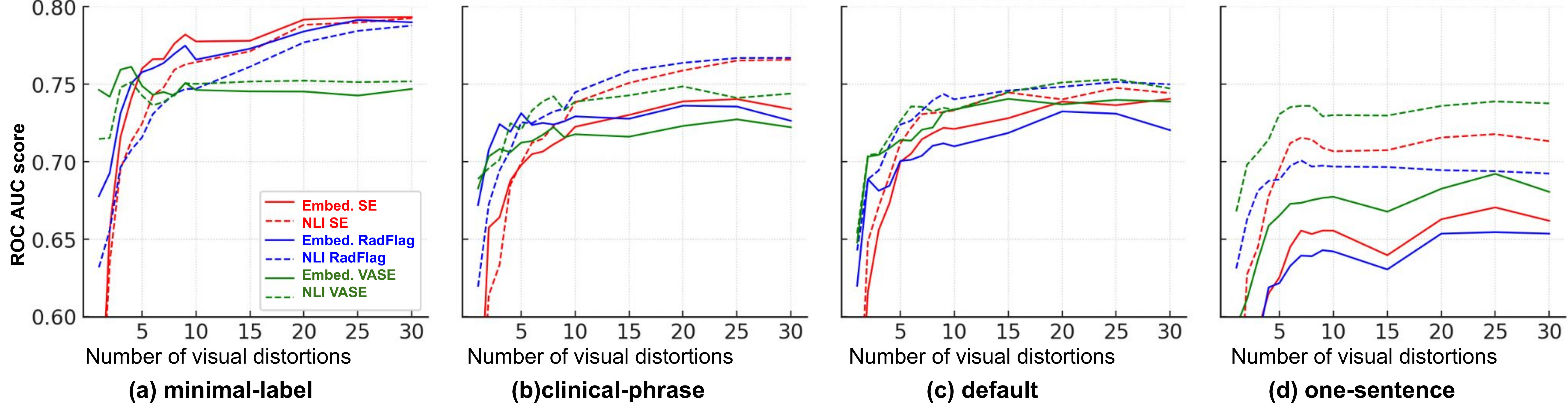}
    \centering
\caption{
Effect of sampling scale and the associated visual perturbations on hallucination-detection performance for {Qwen2.5-VL-7B-Instruct} evaluated on the {VQA-RAD} dataset.
ROC AUC scores are reported for three methods—Semantic Entropy (SE), RadFlag, and Vision-Amplified Semantic Entropy (VASE)—evaluated {across four answer-length prompt configurations}, including minimal-label, clinical-phrase, default, and one-sentence.
Solid lines denote embedding-based clustering, whereas dashed lines denote NLI-based clustering.
}
\label{fig:auc_vs_distortion}   
\end{figure*}

\begin{table*}[h!]
\footnotesize
\centering
\caption{
Hallucination detection performance (\textbf{ROC AUC}) of \textbf{Qwen2.5-VL-7B-Instruct} on the \textbf{VQA-RAD} dataset across varying sampling scales—where each configuration uses an equal number of clean high-temperature samples and visually perturbed counterparts—for both embedding- and NLI-based clustering approaches. Each block corresponds to one answer-length prompt configuration—minimal-label, clinical-phrase, default and one-sentence. Rows list the three evaluation metrics (SE, RadFlag, and VASE), and columns indicate the number of clean and noisy samples per configuration (1–30). Boldfaced entries mark the best-performing configuration for each method within a given prompt type (based on raw, pre-rounded values).
}
\label{tab:sample_variations}
\begin{tabular}{lcccccccccccccc}
\toprule
 & \multicolumn{14}{c}{\textbf{Clustering by Embedding (our method)}} \\
\cmidrule(lr){2-15}
\textbf{\#Samples} & \textbf{1} & \textbf{2} & \textbf{3} & \textbf{4} & \textbf{5} & \textbf{6} & \textbf{7} & \textbf{8} & \textbf{9} & \textbf{10} & \textbf{15} & \textbf{20} & \textbf{25} & \textbf{30} \\
\midrule
\multicolumn{15}{l}{\textbf{minimal-label}} \\
SE & 0.50 & 0.66 & 0.72 & 0.74 & 0.76 & 0.77 & 0.77 & 0.78 & 0.78 & 0.78 & 0.78 & 0.79 & 0.79 & \textbf{0.79} \\
RadFlag & 0.68 & 0.69 & 0.73 & 0.75 & 0.76 & 0.76 & 0.76 & 0.77 & 0.77 & 0.77 & 0.77 & 0.78 & \textbf{0.79} & 0.79 \\
VASE & 0.75 & 0.74 & 0.76 & \textbf{0.76} & 0.75 & 0.74 & 0.74 & 0.74 & 0.75 & 0.75 & 0.75 & 0.75 & 0.74 & 0.75 \\
\midrule
\multicolumn{15}{l}{\textbf{clinical-phrase}} \\
SE & 0.50 & 0.66 & 0.66 & 0.69 & 0.70 & 0.70 & 0.71 & 0.71 & 0.72 & 0.72 & 0.73 & 0.74 & \textbf{0.74} & 0.73 \\
RadFlag & 0.67 & 0.71 & 0.72 & 0.72 & 0.73 & 0.72 & 0.73 & 0.72 & 0.73 & 0.73 & 0.73 & \textbf{0.74} & 0.74 & 0.73 \\
VASE & 0.68 & 0.70 & 0.71 & 0.71 & 0.71 & 0.71 & 0.72 & 0.72 & 0.72 & 0.72 & 0.72 & 0.72 & \textbf{0.73} & 0.72 \\
\midrule
\multicolumn{15}{l}{\textbf{default}} \\
SE & 0.50 & 0.62 & 0.66 & 0.67 & 0.70 & 0.71 & 0.71 & 0.72 & 0.72 & 0.72 & 0.73 & 0.74 & 0.74 & \textbf{0.74} \\
RadFlag & 0.62 & 0.69 & 0.68 & 0.68 & 0.70 & 0.70 & 0.70 & 0.71 & 0.71 & 0.71 & 0.72 & \textbf{0.73} & 0.73 & 0.72 \\
VASE & 0.65 & 0.70 & 0.70 & 0.71 & 0.71 & 0.71 & 0.72 & 0.72 & 0.73 & 0.73 & \textbf{0.74} & 0.74 & 0.74 & 0.74 \\
\midrule
\multicolumn{15}{l}{\textbf{one-sentence}} \\
SE & 0.50 & 0.58 & 0.59 & 0.62 & 0.63 & 0.65 & 0.66 & 0.65 & 0.66 & 0.66 & 0.64 & 0.66 & \textbf{0.67} & 0.66 \\
RadFlag & 0.55 & 0.58 & 0.58 & 0.62 & 0.62 & 0.63 & 0.64 & 0.64 & 0.64 & 0.64 & 0.63 & 0.65 & \textbf{0.65} & 0.65 \\
VASE & 0.59 & 0.61 & 0.64 & 0.66 & 0.67 & 0.67 & 0.67 & 0.68 & 0.68 & 0.68 & 0.67 & 0.68 & \textbf{0.69} & 0.68 \\
\midrule
\\[-1em]
 & \multicolumn{14}{c}{\textbf{Clustering by NLI}} \\
\cmidrule(lr){2-15}
\textbf{\#Samples} & \textbf{1} & \textbf{2} & \textbf{3} & \textbf{4} & \textbf{5} & \textbf{6} & \textbf{7} & \textbf{8} & \textbf{9} & \textbf{10} & \textbf{15} & \textbf{20} & \textbf{25} & \textbf{30} \\
\midrule
\multicolumn{15}{l}{\textbf{minimal-label}} \\
SE & 0.50 & 0.63 & 0.70 & 0.71 & 0.72 & 0.74 & 0.75 & 0.76 & 0.76 & 0.76 & 0.77 & 0.79 & 0.79 & \textbf{0.79} \\
RadFlag & 0.63 & 0.66 & 0.70 & 0.71 & 0.72 & 0.73 & 0.74 & 0.74 & 0.75 & 0.75 & 0.76 & 0.78 & 0.78 & \textbf{0.79} \\
VASE & 0.71 & 0.72 & 0.75 & 0.75 & 0.74 & 0.74 & 0.74 & 0.74 & 0.75 & 0.75 & 0.75 & \textbf{0.75} & 0.75 & 0.75 \\
\midrule
\multicolumn{15}{l}{\textbf{clinical-phrase}} \\
SE & 0.50 & 0.61 & 0.63 & 0.69 & 0.70 & 0.71 & 0.71 & 0.72 & 0.73 & 0.74 & 0.75 & 0.76 & 0.77 & \textbf{0.77} \\
RadFlag & 0.62 & 0.67 & 0.69 & 0.71 & 0.73 & 0.72 & 0.73 & 0.73 & 0.73 & 0.74 & 0.76 & 0.76 & 0.77 & \textbf{0.77} \\
VASE & 0.69 & 0.70 & 0.70 & 0.72 & 0.72 & 0.73 & 0.74 & 0.74 & 0.73 & 0.74 & 0.74 & \textbf{0.75} & 0.74 & 0.74 \\
\midrule
\multicolumn{15}{l}{\textbf{default}} \\
SE & 0.50 & 0.65 & 0.67 & 0.69 & 0.71 & 0.72 & 0.73 & 0.73 & 0.73 & 0.73 & 0.74 & 0.74 & \textbf{0.75} & 0.74 \\
RadFlag & 0.64 & 0.69 & 0.69 & 0.71 & 0.72 & 0.73 & 0.73 & 0.74 & 0.74 & 0.74 & 0.75 & 0.75 & \textbf{0.75} & 0.75 \\
VASE & 0.65 & 0.70 & 0.71 & 0.72 & 0.73 & 0.74 & 0.74 & 0.73 & 0.73 & 0.73 & 0.74 & 0.75 & \textbf{0.75} & 0.75 \\
\midrule
\multicolumn{15}{l}{\textbf{one-sentence}} \\
SE & 0.50 & 0.63 & 0.64 & 0.68 & 0.70 & 0.71 & 0.72 & 0.71 & 0.71 & 0.71 & 0.71 & 0.72 & \textbf{0.72} & 0.71 \\
RadFlag & 0.63 & 0.66 & 0.68 & 0.69 & 0.69 & 0.70 & \textbf{0.70} & 0.70 & 0.70 & 0.70 & 0.70 & 0.69 & 0.69 & 0.69 \\
VASE & 0.67 & 0.70 & 0.71 & 0.71 & 0.73 & 0.74 & 0.74 & 0.74 & 0.73 & 0.73 & 0.73 & 0.74 & \textbf{0.74} & 0.74 \\
\bottomrule
\end{tabular}
\end{table*}

Table~\ref{tab:sample_variations} and Figure~\ref{fig:auc_vs_distortion} study Qwen2.5-VL-7B-Instruct on VQA-RAD as we vary the sampling scale $n$ (equal numbers of clean and noisy high-temperature samples).

\subsubsection{Extreme low budget ($n{=}1$)}
The metrics separate clearly: SE collapses to chance (AUC $=0.50$ for all prompts and clustering methods), RadFlag remains moderately predictive (AUC $\approx 0.55$–$0.68$ across prompts), and VASE is already strong (AUC $\approx 0.59$–$0.75$) even with a single clean–noisy pair.
At $n{=}1$, VASE is the best choice if distortions are available, RadFlag offers a cheap clean-only fallback, and SE is not yet usable.

\subsubsection{Moderate budgets ($n{\approx}10$--15)}
All metrics improve sharply as $n$ increases from 2 to $\sim 10$.
Beyond $n{\approx}10$–15, AUC gains are typically $\leq 0.01$–$0.02$ for all prompts and clustering methods.
This identifies $n{\approx}10$–15 as a practical sweet spot: it recovers most of the available signal without the extra cost of $n{=}25$–30.

\subsubsection{Prompt and metric interactions}
For both NLI and embedding clustering, SE and RadFlag achieve their highest AUCs under \textbf{minimal-label} prompts (e.g., NLI: SE and RadFlag $\approx 0.79$ at $n{\ge}20$), with clinical-phrase and default settings slightly lower and one-sentence responses clearly hardest.
VASE remains robust across all prompt types: it is strong at $n{=}1$ for minimal-label outputs and improves steadily for longer prompts as $n$ increases.

\subsubsection{Clustering and sampling}
Embedding- and NLI-based clustering show similar saturation with $n$ but diverge in specific regimes: for minimal-label and clinical-phrase prompts, both strategies converge to nearly identical AUCs once $n{\gtrsim}10$, whereas for the most verbose one-sentence setting, NLI consistently outperforms embeddings (e.g., SE at large $n$: $\approx 0.72$ NLI vs.\ $\approx 0.67$ embedding; VASE $\approx 0.74$ vs.\ $\approx 0.69$).

Small non-monotonic bumps (e.g., VASE minimal-label / embedding around $n{=}4$) are rare and numerically small.

\subsubsection{Practical recommendations}
For models like Qwen2.5-VL-7B-Instruct, at $n{=}1$–2 it is best to rely on VASE when distortions are available and fall back to RadFlag otherwise, while SE should be avoided.
At $n{\approx}10$–15, SE, RadFlag, and VASE all perform strongly, making this sampling budget the best cost–performance trade-off.
When answers become verbose, VASE paired with NLI clustering remains the most reliable configuration.

\subsection{Summary of Main Findings}
\label{subsec:results_summary}

Across all tables and configurations, Qwen2.5-VL produces the most informative uncertainty structure, whereas Med-Gemma lags behind.
KvasirVQA-x1 remains consistently easier for hallucination detection than VQA-RAD.
Default and clinical-phrase prompts are the safest choices, minimal-label prompts are powerful for strong models but fragile for weaker ones, and one-sentence prompts represent the hardest regime.
NLI and embedding clustering are complementary: NLI is better for verbose and underspecified answers (especially for LLaVA-Med, KvasirVQA-x1, and one-sentence prompts), while embeddings excel for compact, label-like outputs.
VASE emerges as the most robust metric overall once at least one clean–noisy pair is available, though SE and RadFlag remain competitive and useful in specific budget regimes.

\section{Discussion}
\label{sec:discussion}

\Gls{hedge} exposes how model architecture, dataset design, prompt choices, clustering strategy, and sampling scale jointly control hallucination detectability.
Here, we highlight the main factors and practical lessons.

\subsection{Architecture and Data Matter}
\label{subsec:architecture}

The strong and stable performance of Qwen2.5-VL compared to LLaVA-Med and Med-Gemma indicates that hallucination detection quality is not just about the metric, but also about the underlying vision–language representation.
Dense, unified fusion with more visual tokens (as in Qwen2.5-VL) produces smoother embedding manifolds and more informative log-probability patterns than cropped or heavily bottlenecked pipelines.

Dataset geometry amplifies this effect.
KvasirVQA-x1’s relatively homogeneous endoscopy images and localized lesions make hallucinations easier to separate than VQA-RAD’s heterogeneous radiology images.
Absolute AUC values therefore reflect both semantic difficulty and low-level imaging characteristics.

\subsection{Sampling Scale and Robustness Budget}
\label{subsec:discussion_sampling}

Sampling scale is a critical knob in \gls{hedge}.

\begin{itemize}
    \item At $n{=}1$, VASE already delivers strong performance by exploiting a single clean–noisy pair, while RadFlag is usable and SE fails.
    \item At $n{\approx}10$–15, all metrics stabilize; additional samples rarely add more than $0.01$–$0.02$ AUC.
\end{itemize}

Embedding-based clustering continues to benefit up to this sweet spot, indicating that it integrates redundant evidence well.
NLI-based clustering saturates earlier and is more sensitive to paraphrastic noise at large $n$.
In practice, $n{\approx}10$–15 with equal clean and noisy samples is a good default budget.

In our design, increasing $n$ also increases the number of distorted variants, so sampling scale and perturbation budget are coupled.
Future work should decouple these factors to separately measure the impact of more samples versus stronger perturbations.

\subsection{Choosing a Metric and Clustering Strategy}
\label{subsec:discussion_metrics}
Within the \gls{hedge} framework, our results suggest a simple decision guide:
\begin{itemize}
    \item \textbf{Metric choice.}
    If distortions are available, use VASE as the primary metric.
    For label-style outputs and moderate $n$, SE and RadFlag are competitive and cheaper.
    At $n{=}1$, VASE and RadFlag are the only viable options.
    \item \textbf{Clustering choice.}
    Use NLI when answers are long or underspecified (e.g., one-sentence, default with rich language, or yes/no labels).
    Use embeddings when answers are short and label-like and when a tuned $\tau$ is acceptable.
\end{itemize}

There is no single best combination, but “VASE + NLI” for verbose answers and “SE/RadFlag + embeddings” for minimal-label regimes emerge as strong anchors.

\subsection{Adjudication and Reliability}
\label{subsec:adjudicator}

Ground-truth hallucination labels in our experiments come from a single LLM adjudicator with few-shot calibration.
This is efficient but introduces potential bias and provides no direct measure of label uncertainty.
For future benchmarks, we recommend
(i) using multiple independent adjudicators (LLMs and/or experts),
(ii) reporting agreement statistics, and
(iii) releasing adjudication rationales to enable case-level error analysis.

\subsection{Future Work}
\label{subsec:future}

Several extensions follow naturally:

\begin{itemize}
    \item Standardize image preprocessing across models and datasets to disentangle metric, model, and data effects.
    \item Report confidence intervals and paired significance tests alongside AUC.
    \item Explore hybrid clustering that fuses entailment and embeddings in a single graph.
    \item Extend \gls{hedge} to longer clinical reports, non-medical domains, and multilingual settings.
\end{itemize}

\subsection{Key Takeaways}
\label{subsec:takeaway}

\Gls{hedge} shows that hallucination detection is not a single scalar problem but an interaction of architecture, dataset, prompt, clustering, and sampling choices.
Within this space, a few clear lessons emerge:

\begin{itemize}
    \item Strong VLMs (like Qwen2.5-VL) expose more informative uncertainty structure than weaker ones.
    \item Default and clinical-phrase prompts are the safest regimes; one-sentence is the most fragile.
    \item A moderate sampling budget ($n{\approx}10$–15) with equal clean and noisy counts is usually enough.
    \item VASE provides the most robust scores across conditions, but SE and RadFlag remain valuable, especially in tight compute regimes.
\end{itemize}

These patterns give concrete, actionable guidance for deploying hallucination detection in real multimodal systems and for designing stronger future benchmarks.

\section{Conclusion}
\label{sec:conclusion}

We introduced \gls{hedge}, a modular framework for hallucination detection in vision–language models that unifies sampling, semantic clustering, and robustness-based scoring. Across two medical VQA benchmarks and three VLM architectures, our study shows that hallucination detectability is driven less by any single metric and more by the interaction between sampling scale, prompt design, clustering strategy, and the underlying vision–language representation.

Our experiments reveal three consistent insights.  
(1) Stronger architectures with dense, unified fusion (e.g., Qwen2.5-VL) produce uncertainty patterns that are substantially easier to separate than cropped or bottlenecked pipelines.  
(2) Prompting and sampling matter: concise but expressive answers and a moderate sampling budget ($n{\approx}10$–15 clean–noisy pairs) provide the most reliable signal.  
(3) VASE is the most robust metric overall, while SE and RadFlag remain competitive under label-style prompts and low-cost settings; NLI and embedding clustering excel in complementary regimes.

Rather than treating hallucination detection as a static leaderboard, \gls{hedge} reframes it as a structured robustness analysis shaped by model architecture, dataset geometry, and semantic variability. The framework provides a reproducible, compute-aware foundation for evaluating future multimodal systems and opens a path toward standardized, transparent benchmarking of hallucination robustness.

\paragraph*{Acknowledgments}
This work was partly funded by the Research Council of Norway, project number 346671 (AI-Storyteller), and has benefited from the Experimental Infrastructure for Exploration of Exascale Computing (eX3), which is financially supported by the Research Council of Norway under contract 270053.

\paragraph*{Declaration on Generative AI}
During the preparation of this work, the author(s) used {GPT~-5.1} in order to perform \textit{grammar and spelling checks, paraphrasing and rewording, and improving the writing style}. After using this tool/service, the author(s) reviewed and edited the content as needed and take full responsibility for the publication’s content.

\bibliographystyle{ACM-Reference-Format}
\bibliography{references}


\begin{thebibliography}{95}


\ifx \showCODEN    \undefined \def \showCODEN     #1{\unskip}     \fi
\ifx \showISBNx    \undefined \def \showISBNx     #1{\unskip}     \fi
\ifx \showISBNxiii \undefined \def \showISBNxiii  #1{\unskip}     \fi
\ifx \showISSN     \undefined \def \showISSN      #1{\unskip}     \fi
\ifx \showLCCN     \undefined \def \showLCCN      #1{\unskip}     \fi
\ifx \shownote     \undefined \def \shownote      #1{#1}          \fi
\ifx \showarticletitle \undefined \def \showarticletitle #1{#1}   \fi
\ifx \showURL      \undefined \def \showURL       {\relax}        \fi
\providecommand\bibfield[2]{#2}
\providecommand\bibinfo[2]{#2}
\providecommand\natexlab[1]{#1}
\providecommand\showeprint[2][]{arXiv:#2}

\bibitem[Akiba et~al\mbox{.}(2019)]%
        {Akiba2019Jul}
\bibfield{author}{\bibinfo{person}{Takuya Akiba}, \bibinfo{person}{Shotaro Sano}, \bibinfo{person}{Toshihiko Yanase}, {et~al\mbox{.}}} \bibinfo{year}{2019}\natexlab{}.
\newblock \showarticletitle{{Optuna: A Next-generation Hyperparameter Optimization Framework}}.
\newblock \bibinfo{journal}{\emph{ArXiv e-prints}} (\bibinfo{date}{July} \bibinfo{year}{2019}).
\newblock
\href{https://doi.org/10.48550/arXiv.1907.10902}{doi:\nolinkurl{10.48550/arXiv.1907.10902}}


\bibitem[Bahng et~al\mbox{.}(2025)]%
        {CycleConsistency2025}
\bibfield{author}{\bibinfo{person}{Hyojin Bahng}, \bibinfo{person}{Caroline Chan}, \bibinfo{person}{Fredo Durand}, {et~al\mbox{.}}} \bibinfo{year}{2025}\natexlab{}.
\newblock \showarticletitle{{Cycle Consistency as Reward: Learning Image-Text Alignment without Human Preferences}}.
\newblock \bibinfo{journal}{\emph{ArXiv e-prints}} (\bibinfo{date}{June} \bibinfo{year}{2025}).
\newblock
\href{https://doi.org/10.48550/arXiv.2506.02095}{doi:\nolinkurl{10.48550/arXiv.2506.02095}}


\bibitem[Bai et~al\mbox{.}(2025)]%
        {Qwen25vl}
\bibfield{author}{\bibinfo{person}{Shuai Bai}, \bibinfo{person}{Keqin Chen}, \bibinfo{person}{Xuejing Liu}, {et~al\mbox{.}}} \bibinfo{year}{2025}\natexlab{}.
\newblock \showarticletitle{{Qwen2.5-VL Technical Report}}.
\newblock \bibinfo{journal}{\emph{ArXiv e-prints}} (\bibinfo{date}{Feb.} \bibinfo{year}{2025}).
\newblock
\href{https://doi.org/10.48550/arXiv.2502.13923}{doi:\nolinkurl{10.48550/arXiv.2502.13923}}


\bibitem[Bang et~al\mbox{.}(2025)]%
        {Bang2025HalluLens}
\bibfield{author}{\bibinfo{person}{Yejin Bang}, \bibinfo{person}{Ziwei Ji}, \bibinfo{person}{Alan Schelten}, \bibinfo{person}{Anthony Hartshorn}, \bibinfo{person}{Tara Fowler}, \bibinfo{person}{Cheng Zhang}, \bibinfo{person}{Nicola Cancedda}, {and} \bibinfo{person}{Pascale Fung}.} \bibinfo{year}{2025}\natexlab{}.
\newblock \showarticletitle{{H}allu{L}ens: {LLM} Hallucination Benchmark}. In \bibinfo{booktitle}{\emph{Proceedings of the 63rd Annual Meeting of the Association for Computational Linguistics (Volume 1: Long Papers)}}, \bibfield{editor}{\bibinfo{person}{Wanxiang Che}, \bibinfo{person}{Joyce Nabende}, \bibinfo{person}{Ekaterina Shutova}, {and} \bibinfo{person}{Mohammad~Taher Pilehvar}} (Eds.). \bibinfo{publisher}{Association for Computational Linguistics}, \bibinfo{address}{Vienna, Austria}, \bibinfo{pages}{24128--24156}.
\newblock
\showISBNx{979-8-89176-251-0}
\href{https://doi.org/10.18653/v1/2025.acl-long.1176}{doi:\nolinkurl{10.18653/v1/2025.acl-long.1176}}


\bibitem[Bastola and Choi(2025)]%
        {bastola2025hybrid}
\bibfield{author}{\bibinfo{person}{Deepak Bastola} {and} \bibinfo{person}{Woohyeok Choi}.} \bibinfo{year}{2025}\natexlab{}.
\newblock \showarticletitle{{Hybrid Topic-Semantic Labeling and Graph Embeddings for Unsupervised Legal Document Clustering}}.
\newblock \bibinfo{journal}{\emph{ArXiv e-prints}} (\bibinfo{date}{Aug.} \bibinfo{year}{2025}).
\newblock
\href{https://doi.org/10.48550/arXiv.2509.00990}{doi:\nolinkurl{10.48550/arXiv.2509.00990}}


\bibitem[Ben-Kish et~al\mbox{.}(2023a)]%
        {OpenCHAIR2024}
\bibfield{author}{\bibinfo{person}{Assaf Ben-Kish}, \bibinfo{person}{Moran Yanuka}, \bibinfo{person}{Morris Alper}, {et~al\mbox{.}}} \bibinfo{year}{2023}\natexlab{a}.
\newblock \showarticletitle{{Mitigating Open-Vocabulary Caption Hallucinations}}.
\newblock \bibinfo{journal}{\emph{ArXiv e-prints}} (\bibinfo{date}{Dec.} \bibinfo{year}{2023}).
\newblock
\href{https://doi.org/10.18653/v1/2024.findings-acl.657}{doi:\nolinkurl{10.18653/v1/2024.findings-acl.657}}


\bibitem[Ben-Kish et~al\mbox{.}(2023b)]%
        {MOCHa2024}
\bibfield{author}{\bibinfo{person}{Assaf Ben-Kish}, \bibinfo{person}{Moran Yanuka}, \bibinfo{person}{Morris Alper}, {et~al\mbox{.}}} \bibinfo{year}{2023}\natexlab{b}.
\newblock \showarticletitle{{MOCHa: Multi-Objective Reinforcement Mitigating Caption Hallucinations}}.
\newblock \bibinfo{journal}{\emph{ArXiv e-prints}} (\bibinfo{date}{Dec.} \bibinfo{year}{2023}).
\newblock
\href{https://doi.org/10.48550/arXiv.2312.03631}{doi:\nolinkurl{10.48550/arXiv.2312.03631}}


\bibitem[Berant(2012)]%
        {berant2011global}
\bibfield{author}{\bibinfo{person}{Jonathan Berant}.} \bibinfo{year}{2012}\natexlab{}.
\newblock \emph{\bibinfo{title}{Global Learning of Textual Entailment Graphs}}.
\newblock \bibinfo{thesistype}{Ph.\,D. Dissertation}. \bibinfo{school}{Tel Aviv University}.
\newblock


\bibitem[Bergeron et~al\mbox{.}(2025)]%
        {HalluGuard2025}
\bibfield{author}{\bibinfo{person}{Loris Bergeron}, \bibinfo{person}{Ioana Buhnila}, \bibinfo{person}{J{\ifmmode\acute{e}\else\'{e}\fi}r{\ifmmode\hat{o}\else\^{o}\fi}me Fran{\ifmmode\mbox{\c{c}}\else\c{c}\fi}ois}, {et~al\mbox{.}}} \bibinfo{year}{2025}\natexlab{}.
\newblock \showarticletitle{{HalluGuard: Evidence-Grounded Small Reasoning Models to Mitigate Hallucinations in Retrieval-Augmented Generation}}.
\newblock \bibinfo{journal}{\emph{ArXiv e-prints}} (\bibinfo{date}{Oct.} \bibinfo{year}{2025}).
\newblock
\href{https://doi.org/10.48550/arXiv.2510.00880}{doi:\nolinkurl{10.48550/arXiv.2510.00880}}


\bibitem[Bn et~al\mbox{.}(2025)]%
        {Suhas2025ClinicalNLI}
\bibfield{author}{\bibinfo{person}{Suhas Bn}, \bibinfo{person}{Han-Chin Shing}, \bibinfo{person}{Lei Xu}, {et~al\mbox{.}}} \bibinfo{year}{2025}\natexlab{}.
\newblock \showarticletitle{{Fact-Controlled Diagnosis of Hallucinations in Medical Text Summarization}}.
\newblock \bibinfo{journal}{\emph{ArXiv e-prints}} (\bibinfo{date}{May} \bibinfo{year}{2025}).
\newblock
\href{https://doi.org/10.48550/arXiv.2506.00448}{doi:\nolinkurl{10.48550/arXiv.2506.00448}}


\bibitem[Buslaev et~al\mbox{.}(2018)]%
        {Buslaev2018Sep}
\bibfield{author}{\bibinfo{person}{Alexander Buslaev}, \bibinfo{person}{Alex Parinov}, \bibinfo{person}{Eugene Khvedchenya}, {et~al\mbox{.}}} \bibinfo{year}{2018}\natexlab{}.
\newblock \showarticletitle{{Albumentations: fast and flexible image augmentations}}.
\newblock \bibinfo{journal}{\emph{ArXiv e-prints}} (\bibinfo{date}{Sept.} \bibinfo{year}{2018}).
\newblock
\href{https://doi.org/10.3390/info11020125}{doi:\nolinkurl{10.3390/info11020125}}


\bibitem[Chen et~al\mbox{.}(2024a)]%
        {DiaHalu2024}
\bibfield{author}{\bibinfo{person}{Kedi Chen}, \bibinfo{person}{Qin Chen}, \bibinfo{person}{Jie Zhou}, \bibinfo{person}{He Yishen}, {and} \bibinfo{person}{Liang He}.} \bibinfo{year}{2024}\natexlab{a}.
\newblock \showarticletitle{{D}ia{H}alu: A Dialogue-level Hallucination Evaluation Benchmark for Large Language Models}. In \bibinfo{booktitle}{\emph{Findings of the Association for Computational Linguistics: EMNLP 2024}}, \bibfield{editor}{\bibinfo{person}{Yaser Al-Onaizan}, \bibinfo{person}{Mohit Bansal}, {and} \bibinfo{person}{Yun-Nung Chen}} (Eds.). \bibinfo{publisher}{Association for Computational Linguistics}, \bibinfo{address}{Miami, Florida, USA}, \bibinfo{pages}{9057--9079}.
\newblock
\href{https://doi.org/10.18653/v1/2024.findings-emnlp.529}{doi:\nolinkurl{10.18653/v1/2024.findings-emnlp.529}}


\bibitem[Chen et~al\mbox{.}(2023)]%
        {FACTCHD2024}
\bibfield{author}{\bibinfo{person}{Xiang Chen}, \bibinfo{person}{Duanzheng Song}, \bibinfo{person}{Honghao Gui}, {et~al\mbox{.}}} \bibinfo{year}{2023}\natexlab{}.
\newblock \showarticletitle{{FactCHD: Benchmarking Fact-Conflicting Hallucination Detection}}.
\newblock \bibinfo{journal}{\emph{ArXiv e-prints}} (\bibinfo{date}{Oct.} \bibinfo{year}{2023}).
\newblock
\href{https://doi.org/10.48550/arXiv.2310.12086}{doi:\nolinkurl{10.48550/arXiv.2310.12086}}


\bibitem[Chen et~al\mbox{.}(2024b)]%
        {MHaluBench2024}
\bibfield{author}{\bibinfo{person}{Xiang Chen}, \bibinfo{person}{Chenxi Wang}, \bibinfo{person}{Yida Xue}, {et~al\mbox{.}}} \bibinfo{year}{2024}\natexlab{b}.
\newblock \showarticletitle{{Unified Hallucination Detection for Multimodal Large Language Models}}.
\newblock \bibinfo{journal}{\emph{ACL Anthology}} (\bibinfo{date}{Aug.} \bibinfo{year}{2024}), \bibinfo{pages}{3235--3252}.
\newblock
\href{https://doi.org/10.18653/v1/2024.acl-long.178}{doi:\nolinkurl{10.18653/v1/2024.acl-long.178}}


\bibitem[Chen et~al\mbox{.}(2025)]%
        {MMSurvey2025}
\bibfield{author}{\bibinfo{person}{Zhiyuan Chen}, \bibinfo{person}{Yuecong Min}, \bibinfo{person}{Jie Zhang}, {et~al\mbox{.}}} \bibinfo{year}{2025}\natexlab{}.
\newblock \showarticletitle{{A Survey of Multimodal Hallucination Evaluation and Detection}}.
\newblock \bibinfo{journal}{\emph{ArXiv e-prints}} (\bibinfo{date}{July} \bibinfo{year}{2025}).
\newblock
\href{https://doi.org/10.48550/arXiv.2507.19024}{doi:\nolinkurl{10.48550/arXiv.2507.19024}}


\bibitem[Cheng et~al\mbox{.}(2025)]%
        {CapArena2024}
\bibfield{author}{\bibinfo{person}{Kanzhi Cheng}, \bibinfo{person}{Wenpo Song}, \bibinfo{person}{Jiaxin Fan}, {et~al\mbox{.}}} \bibinfo{year}{2025}\natexlab{}.
\newblock \showarticletitle{{CapArena: Benchmarking and Analyzing Detailed Image Captioning in the LLM Era}}.
\newblock \bibinfo{journal}{\emph{ArXiv e-prints}} (\bibinfo{date}{March} \bibinfo{year}{2025}).
\newblock
\href{https://doi.org/10.48550/arXiv.2503.12329}{doi:\nolinkurl{10.48550/arXiv.2503.12329}}


\bibitem[Dagan et~al\mbox{.}(2006)]%
        {dagan2006pascal}
\bibfield{author}{\bibinfo{person}{Ido Dagan}, \bibinfo{person}{Oren Glickman}, {and} \bibinfo{person}{Bernardo Magnini}.} \bibinfo{year}{2006}\natexlab{}.
\newblock \showarticletitle{The PASCAL Recognising Textual Entailment Challenge}.
\newblock In \bibinfo{booktitle}{\emph{Machine Learning Challenges. Evaluating Predictive Uncertainty, Visual Object Classification, and Recognising Textual Entailment}}. \bibinfo{series}{Lecture Notes in Computer Science}, Vol.~\bibinfo{volume}{3944}. \bibinfo{publisher}{Springer}, \bibinfo{pages}{177--190}.
\newblock
\href{https://doi.org/10.1007/11736790_9}{doi:\nolinkurl{10.1007/11736790_9}}


\bibitem[Devlin et~al\mbox{.}(2018)]%
        {devlin2018bert}
\bibfield{author}{\bibinfo{person}{Jacob Devlin}, \bibinfo{person}{Ming-Wei Chang}, \bibinfo{person}{Kenton Lee}, {et~al\mbox{.}}} \bibinfo{year}{2018}\natexlab{}.
\newblock \showarticletitle{{BERT: Pre-training of Deep Bidirectional Transformers for Language Understanding}}.
\newblock \bibinfo{journal}{\emph{ArXiv e-prints}} (\bibinfo{date}{Oct.} \bibinfo{year}{2018}).
\newblock
\href{https://doi.org/10.48550/arXiv.1810.04805}{doi:\nolinkurl{10.48550/arXiv.1810.04805}}


\bibitem[Dziri et~al\mbox{.}(2022)]%
        {FaithDial2022}
\bibfield{author}{\bibinfo{person}{Nouha Dziri}, \bibinfo{person}{Ehsan Kamalloo}, \bibinfo{person}{Sivan Milton}, \bibinfo{person}{Osmar Zaiane}, \bibinfo{person}{Mo Yu}, \bibinfo{person}{Edoardo~M. Ponti}, {and} \bibinfo{person}{Siva Reddy}.} \bibinfo{year}{2022}\natexlab{}.
\newblock \showarticletitle{{F}aith{D}ial: A Faithful Benchmark for Information-Seeking Dialogue}.
\newblock \bibinfo{journal}{\emph{Transactions of the Association for Computational Linguistics}}  \bibinfo{volume}{10} (\bibinfo{year}{2022}), \bibinfo{pages}{1473--1490}.
\newblock
\href{https://doi.org/10.1162/tacl_a_00529}{doi:\nolinkurl{10.1162/tacl_a_00529}}


\bibitem[Es et~al\mbox{.}(2024)]%
        {RAGAS2024}
\bibfield{author}{\bibinfo{person}{Shahul Es}, \bibinfo{person}{Jithin James}, \bibinfo{person}{Luis Espinosa~Anke}, {and} \bibinfo{person}{Steven Schockaert}.} \bibinfo{year}{2024}\natexlab{}.
\newblock \showarticletitle{{RAGA}s: Automated Evaluation of Retrieval Augmented Generation}. In \bibinfo{booktitle}{\emph{Proceedings of the 18th Conference of the European Chapter of the Association for Computational Linguistics: System Demonstrations}}, \bibfield{editor}{\bibinfo{person}{Nikolaos Aletras} {and} \bibinfo{person}{Orphee De~Clercq}} (Eds.). \bibinfo{publisher}{Association for Computational Linguistics}, \bibinfo{address}{St. Julians, Malta}, \bibinfo{pages}{150--158}.
\newblock
\href{https://doi.org/10.18653/v1/2024.eacl-demo.16}{doi:\nolinkurl{10.18653/v1/2024.eacl-demo.16}}


\bibitem[Fabian(2011)]%
        {fabian2011scikit}
\bibfield{author}{\bibinfo{person}{Pedregosa Fabian}.} \bibinfo{year}{2011}\natexlab{}.
\newblock \showarticletitle{Scikit-learn: Machine learning in Python}.
\newblock \bibinfo{journal}{\emph{Journal of machine learning research 12}} (\bibinfo{year}{2011}), \bibinfo{pages}{2825}.
\newblock


\bibitem[Falke et~al\mbox{.}(2019)]%
        {Falke2019NLI}
\bibfield{author}{\bibinfo{person}{Tobias Falke}, \bibinfo{person}{Leonardo F.~R. Ribeiro}, \bibinfo{person}{Prasetya~Ajie Utama}, \bibinfo{person}{Ido Dagan}, {and} \bibinfo{person}{Iryna Gurevych}.} \bibinfo{year}{2019}\natexlab{}.
\newblock \showarticletitle{Ranking Generated Summaries by Correctness: An Interesting but Challenging Application for Natural Language Inference}. In \bibinfo{booktitle}{\emph{Proceedings of the 57th Annual Meeting of the Association for Computational Linguistics}}, \bibfield{editor}{\bibinfo{person}{Anna Korhonen}, \bibinfo{person}{David Traum}, {and} \bibinfo{person}{Llu{\'i}s M{\`a}rquez}} (Eds.). \bibinfo{publisher}{Association for Computational Linguistics}, \bibinfo{address}{Florence, Italy}, \bibinfo{pages}{2214--2220}.
\newblock
\href{https://doi.org/10.18653/v1/P19-1213}{doi:\nolinkurl{10.18653/v1/P19-1213}}


\bibitem[Farquhar et~al\mbox{.}(2024)]%
        {SE2024}
\bibfield{author}{\bibinfo{person}{Sebastian Farquhar}, \bibinfo{person}{Jannik Kossen}, \bibinfo{person}{Lorenz Kuhn}, {and} \bibinfo{person}{Yarin Gal}.} \bibinfo{year}{2024}\natexlab{}.
\newblock \showarticletitle{{Detecting hallucinations in large language models using semantic entropy}}.
\newblock \bibinfo{journal}{\emph{Nature}} \bibinfo{volume}{630}, \bibinfo{number}{8017} (\bibinfo{date}{June} \bibinfo{year}{2024}), \bibinfo{pages}{625--630}.
\newblock
\showISSN{1476-4687}
\href{https://doi.org/10.1038/s41586-024-07421-0}{doi:\nolinkurl{10.1038/s41586-024-07421-0}}


\bibitem[Fawcett(2006)]%
        {fawcett2006ROC}
\bibfield{author}{\bibinfo{person}{Tom Fawcett}.} \bibinfo{year}{2006}\natexlab{}.
\newblock \showarticletitle{An introduction to ROC analysis}.
\newblock \bibinfo{journal}{\emph{Pattern recognition letters}} \bibinfo{volume}{27}, \bibinfo{number}{8} (\bibinfo{year}{2006}), \bibinfo{pages}{861--874}.
\newblock


\bibitem[Gao et~al\mbox{.}(2021)]%
        {gao2021simcse}
\bibfield{author}{\bibinfo{person}{Tianyu Gao}, \bibinfo{person}{Xingcheng Yao}, {and} \bibinfo{person}{Danqi Chen}.} \bibinfo{year}{2021}\natexlab{}.
\newblock \showarticletitle{SimCSE: Simple Contrastive Learning of Sentence Embeddings}. In \bibinfo{booktitle}{\emph{Proceedings of the 2021 Conference on Empirical Methods in Natural Language Processing (EMNLP)}}. Association for Computational Linguistics, \bibinfo{address}{Punta Cana, Dominican Republic}, \bibinfo{pages}{6894--6910}.
\newblock
\href{https://doi.org/10.18653/v1/2021.emnlp-main.552}{doi:\nolinkurl{10.18653/v1/2021.emnlp-main.552}}


\bibitem[Gao et~al\mbox{.}(2024)]%
        {gao2024spuq}
\bibfield{author}{\bibinfo{person}{Xiang Gao}, \bibinfo{person}{Jiaxin Zhang}, \bibinfo{person}{Lalla Mouatadid}, {and} \bibinfo{person}{Kamalika Das}.} \bibinfo{year}{2024}\natexlab{}.
\newblock \showarticletitle{SPUQ: Perturbation-Based Uncertainty Quantification for Large Language Models}. In \bibinfo{booktitle}{\emph{Proceedings of EACL (Long Papers)}}. \bibinfo{pages}{2336--2346}.
\newblock
\href{https://doi.org/10.18653/v1/2024.eacl-long.143}{doi:\nolinkurl{10.18653/v1/2024.eacl-long.143}}


\bibitem[Gautam et~al\mbox{.}(2025)]%
        {gautam2025kvasirvqax1}
\bibfield{author}{\bibinfo{person}{Sushant Gautam}, \bibinfo{person}{Michael~A. Riegler}, {and} \bibinfo{person}{P{\aa}l Halvorsen}.} \bibinfo{year}{2025}\natexlab{}.
\newblock \showarticletitle{{Kvasir-VQA-x1: A Multimodal Dataset for Medical Reasoning and Robust MedVQA in Gastrointestinal Endoscopy}}.
\newblock \bibinfo{journal}{\emph{ArXiv e-prints}} (\bibinfo{date}{June} \bibinfo{year}{2025}).
\newblock
\href{https://doi.org/10.48550/arXiv.2506.09958}{doi:\nolinkurl{10.48550/arXiv.2506.09958}}


\bibitem[Gautam et~al\mbox{.}(2024)]%
        {gautam2024kvasirvqa}
\bibfield{author}{\bibinfo{person}{Sushant Gautam}, \bibinfo{person}{Andrea Storås}, \bibinfo{person}{Cise Midoglu}, \bibinfo{person}{Steven~A. Hicks}, \bibinfo{person}{Vajira Thambawita}, \bibinfo{person}{Pål Halvorsen}, {and} \bibinfo{person}{Michael~A. Riegler}.} \bibinfo{year}{2024}\natexlab{}.
\newblock \showarticletitle{Kvasir-VQA: A Text-Image Pair GI Tract Dataset}. In \bibinfo{booktitle}{\emph{Proceedings of the First International Workshop on Vision-Language Models for Biomedical Applications (VLM4Bio '24)}} (Melbourne, VIC, Australia). \bibinfo{publisher}{ACM}, \bibinfo{pages}{10 pages}.
\newblock
\href{https://doi.org/10.1145/3689096.3689458}{doi:\nolinkurl{10.1145/3689096.3689458}}


\bibitem[Goel et~al\mbox{.}(2025)]%
        {FinchZK2025}
\bibfield{author}{\bibinfo{person}{Aman Goel}, \bibinfo{person}{Daniel Schwartz}, {and} \bibinfo{person}{Yanjun Qi}.} \bibinfo{year}{2025}\natexlab{}.
\newblock \showarticletitle{{Zero-knowledge LLM hallucination detection and mitigation through fine-grained cross-model consistency}}.
\newblock \bibinfo{journal}{\emph{ArXiv e-prints}} (\bibinfo{date}{Aug.} \bibinfo{year}{2025}).
\newblock
\href{https://doi.org/10.48550/arXiv.2508.14314}{doi:\nolinkurl{10.48550/arXiv.2508.14314}}


\bibitem[Gu et~al\mbox{.}(2024)]%
        {JudgeBiasSurvey2024}
\bibfield{author}{\bibinfo{person}{Jiawei Gu}, \bibinfo{person}{Xuhui Jiang}, \bibinfo{person}{Zhichao Shi}, {et~al\mbox{.}}} \bibinfo{year}{2024}\natexlab{}.
\newblock \showarticletitle{{A Survey on LLM-as-a-Judge}}.
\newblock \bibinfo{journal}{\emph{ArXiv e-prints}} (\bibinfo{date}{Nov.} \bibinfo{year}{2024}).
\newblock
\href{https://doi.org/10.48550/arXiv.2411.15594}{doi:\nolinkurl{10.48550/arXiv.2411.15594}}


\bibitem[Guan et~al\mbox{.}(2023)]%
        {HallusionBench2023}
\bibfield{author}{\bibinfo{person}{Tianrui Guan}, \bibinfo{person}{Fuxiao Liu}, \bibinfo{person}{Xiyang Wu}, {et~al\mbox{.}}} \bibinfo{year}{2023}\natexlab{}.
\newblock \showarticletitle{{HallusionBench: An Advanced Diagnostic Suite for Entangled Language Hallucination and Visual Illusion in Large Vision-Language Models}}.
\newblock \bibinfo{journal}{\emph{ArXiv e-prints}} (\bibinfo{date}{Oct.} \bibinfo{year}{2023}).
\newblock
\href{https://doi.org/10.48550/arXiv.2310.14566}{doi:\nolinkurl{10.48550/arXiv.2310.14566}}


\bibitem[He et~al\mbox{.}(2020)]%
        {DeBERTaHe2020Jun}
\bibfield{author}{\bibinfo{person}{Pengcheng He}, \bibinfo{person}{Xiaodong Liu}, \bibinfo{person}{Jianfeng Gao}, {et~al\mbox{.}}} \bibinfo{year}{2020}\natexlab{}.
\newblock \showarticletitle{{DeBERTa: Decoding-enhanced BERT with Disentangled Attention}}.
\newblock \bibinfo{journal}{\emph{ArXiv e-prints}} (\bibinfo{date}{June} \bibinfo{year}{2020}).
\newblock
\href{https://doi.org/10.48550/arXiv.2006.03654}{doi:\nolinkurl{10.48550/arXiv.2006.03654}}


\bibitem[Honovich et~al\mbox{.}(2022)]%
        {TRUE2021}
\bibfield{author}{\bibinfo{person}{Or Honovich}, \bibinfo{person}{Roee Aharoni}, \bibinfo{person}{Jonathan Herzig}, {et~al\mbox{.}}} \bibinfo{year}{2022}\natexlab{}.
\newblock \showarticletitle{{TRUE: Re-evaluating Factual Consistency Evaluation}}.
\newblock \bibinfo{journal}{\emph{ArXiv e-prints}} (\bibinfo{date}{April} \bibinfo{year}{2022}).
\newblock
\href{https://doi.org/10.48550/arXiv.2204.04991}{doi:\nolinkurl{10.48550/arXiv.2204.04991}}


\bibitem[Hu et~al\mbox{.}(2023)]%
        {MIMICDiffVQA2023}
\bibfield{author}{\bibinfo{person}{Xinyue Hu}, \bibinfo{person}{Lin Gu}, \bibinfo{person}{Qiyuan An}, {et~al\mbox{.}}} \bibinfo{year}{2023}\natexlab{}.
\newblock \showarticletitle{{Expert Knowledge-Aware Image Difference Graph Representation Learning for Difference-Aware Medical Visual Question Answering}}.
\newblock In \bibinfo{booktitle}{\emph{{ACM Conferences}}}. \bibinfo{publisher}{Association for Computing Machinery}, \bibinfo{address}{New York, NY, USA}, \bibinfo{pages}{4156--4165}.
\newblock
\href{https://doi.org/10.1145/3580305.3599819}{doi:\nolinkurl{10.1145/3580305.3599819}}


\bibitem[Huang et~al\mbox{.}(2024)]%
        {AttributionSurvey2024}
\bibfield{author}{\bibinfo{person}{Baixiang Huang}, \bibinfo{person}{Canyu Chen}, {and} \bibinfo{person}{Kai Shu}.} \bibinfo{year}{2024}\natexlab{}.
\newblock \showarticletitle{{Authorship Attribution in the Era of LLMs: Problems, Methodologies, and Challenges}}.
\newblock \bibinfo{journal}{\emph{ArXiv e-prints}} (\bibinfo{date}{Aug.} \bibinfo{year}{2024}).
\newblock
\href{https://doi.org/10.48550/arXiv.2408.08946}{doi:\nolinkurl{10.48550/arXiv.2408.08946}}


\bibitem[Huang et~al\mbox{.}(2025)]%
        {ACMSurvey2025}
\bibfield{author}{\bibinfo{person}{Lei Huang}, \bibinfo{person}{Weijiang Yu}, \bibinfo{person}{Weitao Ma}, {et~al\mbox{.}}} \bibinfo{year}{2025}\natexlab{}.
\newblock \showarticletitle{{A Survey on Hallucination in Large Language Models: Principles, Taxonomy, Challenges, and Open Questions}}.
\newblock \bibinfo{journal}{\emph{ACM Transactions on Information Systems}} \bibinfo{volume}{43}, \bibinfo{number}{2} (\bibinfo{date}{Jan.} \bibinfo{year}{2025}), \bibinfo{pages}{1--55}.
\newblock
\showISSN{1046-8188}
\href{https://doi.org/10.1145/3703155}{doi:\nolinkurl{10.1145/3703155}}


\bibitem[Ifrim and Oncioiu(2025)]%
        {ifrim2025hybridsc}
\bibfield{author}{\bibinfo{person}{Ana-Maria Ifrim} {and} \bibinfo{person}{Ionica Oncioiu}.} \bibinfo{year}{2025}\natexlab{}.
\newblock \showarticletitle{A Hybrid Numerical–Semantic Clustering Algorithm Based on Scalarized Optimization}.
\newblock \bibinfo{journal}{\emph{Algorithms}} \bibinfo{volume}{18}, \bibinfo{number}{10} (\bibinfo{year}{2025}), \bibinfo{pages}{607}.
\newblock
\href{https://doi.org/10.3390/a18100607}{doi:\nolinkurl{10.3390/a18100607}}


\bibitem[Jiang et~al\mbox{.}(2024)]%
        {MiddleLayerProbesCVPR2025}
\bibfield{author}{\bibinfo{person}{Zhangqi Jiang}, \bibinfo{person}{Junkai Chen}, \bibinfo{person}{Beier Zhu}, {et~al\mbox{.}}} \bibinfo{year}{2024}\natexlab{}.
\newblock \showarticletitle{{Devils in Middle Layers of Large Vision-Language Models: Interpreting, Detecting and Mitigating Object Hallucinations via Attention Lens}}.
\newblock \bibinfo{journal}{\emph{ArXiv e-prints}} (\bibinfo{date}{Nov.} \bibinfo{year}{2024}).
\newblock
\href{https://doi.org/10.48550/arXiv.2411.16724}{doi:\nolinkurl{10.48550/arXiv.2411.16724}}


\bibitem[Kapanipathi et~al\mbox{.}(2020)]%
        {kapanipathi2020infusing}
\bibfield{author}{\bibinfo{person}{Pavan Kapanipathi}, \bibinfo{person}{Veronika Thost}, \bibinfo{person}{Siva~Sankalp Patel}, \bibinfo{person}{Spencer Whitehead}, \bibinfo{person}{Ibrahim Abdelaziz}, \bibinfo{person}{Avinash Balakrishnan}, \bibinfo{person}{Maria Chang}, \bibinfo{person}{Kshitij Fadnis}, \bibinfo{person}{Chulaka Gunasekara}, \bibinfo{person}{Bassem Makni}, {et~al\mbox{.}}} \bibinfo{year}{2020}\natexlab{}.
\newblock \showarticletitle{Infusing knowledge into the textual entailment task using graph convolutional networks}. In \bibinfo{booktitle}{\emph{Proceedings of the AAAI conference on artificial intelligence}}, Vol.~\bibinfo{volume}{34}. \bibinfo{pages}{8074--8081}.
\newblock


\bibitem[Kaul et~al\mbox{.}(2024)]%
        {THRONE2024}
\bibfield{author}{\bibinfo{person}{Prannay Kaul}, \bibinfo{person}{Zhizhong Li}, \bibinfo{person}{Hao Yang}, {et~al\mbox{.}}} \bibinfo{year}{2024}\natexlab{}.
\newblock \showarticletitle{{THRONE: An Object-based Hallucination Benchmark for the Free-form Generations of Large Vision-Language Models}}.
\newblock \bibinfo{journal}{\emph{ArXiv e-prints}} (\bibinfo{date}{May} \bibinfo{year}{2024}).
\newblock
\href{https://doi.org/10.48550/arXiv.2405.05256}{doi:\nolinkurl{10.48550/arXiv.2405.05256}}


\bibitem[Kossen et~al\mbox{.}(2024)]%
        {Kossen2024SEP}
\bibfield{author}{\bibinfo{person}{Jannik Kossen}, \bibinfo{person}{Jiatong Han}, \bibinfo{person}{Muhammed Razzak}, {et~al\mbox{.}}} \bibinfo{year}{2024}\natexlab{}.
\newblock \showarticletitle{{Semantic Entropy Probes: Robust and Cheap Hallucination Detection in LLMs}}.
\newblock \bibinfo{journal}{\emph{ArXiv e-prints}} (\bibinfo{date}{June} \bibinfo{year}{2024}).
\newblock
\href{https://doi.org/10.48550/arXiv.2406.15927}{doi:\nolinkurl{10.48550/arXiv.2406.15927}}


\bibitem[Kryscinski et~al\mbox{.}(2020)]%
        {Kryscinski2020FactCC}
\bibfield{author}{\bibinfo{person}{Wojciech Kryscinski}, \bibinfo{person}{Bryan McCann}, \bibinfo{person}{Caiming Xiong}, {and} \bibinfo{person}{Richard Socher}.} \bibinfo{year}{2020}\natexlab{}.
\newblock \showarticletitle{Evaluating the Factual Consistency of Abstractive Text Summarization}. In \bibinfo{booktitle}{\emph{Proceedings of the 2020 Conference on Empirical Methods in Natural Language Processing (EMNLP)}}, \bibfield{editor}{\bibinfo{person}{Bonnie Webber}, \bibinfo{person}{Trevor Cohn}, \bibinfo{person}{Yulan He}, {and} \bibinfo{person}{Yang Liu}} (Eds.). \bibinfo{publisher}{Association for Computational Linguistics}, \bibinfo{address}{Online}, \bibinfo{pages}{9332--9346}.
\newblock
\href{https://doi.org/10.18653/v1/2020.emnlp-main.750}{doi:\nolinkurl{10.18653/v1/2020.emnlp-main.750}}


\bibitem[Laban et~al\mbox{.}(2022)]%
        {Laban2022SummaC}
\bibfield{author}{\bibinfo{person}{Philippe Laban}, \bibinfo{person}{Tobias Schnabel}, \bibinfo{person}{Paul~N. Bennett}, {and} \bibinfo{person}{Marti~A. Hearst}.} \bibinfo{year}{2022}\natexlab{}.
\newblock \showarticletitle{{S}umma{C}: Re-Visiting {NLI}-based Models for Inconsistency Detection in Summarization}.
\newblock \bibinfo{journal}{\emph{Transactions of the Association for Computational Linguistics}}  \bibinfo{volume}{10} (\bibinfo{year}{2022}), \bibinfo{pages}{163--177}.
\newblock
\href{https://doi.org/10.1162/tacl_a_00453}{doi:\nolinkurl{10.1162/tacl_a_00453}}


\bibitem[Lau et~al\mbox{.}(2018)]%
        {lau2018vqarad}
\bibfield{author}{\bibinfo{person}{Jason~J Lau}, \bibinfo{person}{Soumya Gayen}, \bibinfo{person}{Asma Ben~Abacha}, {and} \bibinfo{person}{Dina Demner-Fushman}.} \bibinfo{year}{2018}\natexlab{}.
\newblock \showarticletitle{A dataset of clinically generated visual questions and answers about radiology images}.
\newblock \bibinfo{journal}{\emph{Scientific data}} \bibinfo{volume}{5}, \bibinfo{number}{1} (\bibinfo{year}{2018}), \bibinfo{pages}{1--10}.
\newblock


\bibitem[Li et~al\mbox{.}(2023c)]%
        {LLaVAmed}
\bibfield{author}{\bibinfo{person}{Chunyuan Li}, \bibinfo{person}{Cliff Wong}, \bibinfo{person}{Sheng Zhang}, {et~al\mbox{.}}} \bibinfo{year}{2023}\natexlab{c}.
\newblock \showarticletitle{{LLaVA-med: training a large language-and-vision assistant for biomedicine in one day}}.
\newblock In \bibinfo{booktitle}{\emph{{Guide Proceedings}}}. \bibinfo{publisher}{Curran Associates Inc.}, \bibinfo{pages}{28541--28564}.
\newblock
\href{https://doi.org/10.5555/3666122.3667362}{doi:\nolinkurl{10.5555/3666122.3667362}}


\bibitem[Li et~al\mbox{.}(2024a)]%
        {HaluEval2024}
\bibfield{author}{\bibinfo{person}{Junyi Li}, \bibinfo{person}{Jie Chen}, \bibinfo{person}{Ruiyang Ren}, {et~al\mbox{.}}} \bibinfo{year}{2024}\natexlab{a}.
\newblock \showarticletitle{{The Dawn After the Dark: An Empirical Study on Factuality Hallucination in Large Language Models}}.
\newblock \bibinfo{journal}{\emph{ArXiv e-prints}} (\bibinfo{date}{Jan.} \bibinfo{year}{2024}).
\newblock
\href{https://doi.org/10.48550/arXiv.2401.03205}{doi:\nolinkurl{10.48550/arXiv.2401.03205}}


\bibitem[Li et~al\mbox{.}(2023a)]%
        {HaluEval2022}
\bibfield{author}{\bibinfo{person}{Junyi Li}, \bibinfo{person}{Xiaoxue Cheng}, \bibinfo{person}{Wayne~Xin Zhao}, {et~al\mbox{.}}} \bibinfo{year}{2023}\natexlab{a}.
\newblock \showarticletitle{{HaluEval: A Large-Scale Hallucination Evaluation Benchmark for Large Language Models}}.
\newblock \bibinfo{journal}{\emph{ArXiv e-prints}} (\bibinfo{date}{May} \bibinfo{year}{2023}).
\newblock
\href{https://doi.org/10.48550/arXiv.2305.11747}{doi:\nolinkurl{10.48550/arXiv.2305.11747}}


\bibitem[Li et~al\mbox{.}(2025b)]%
        {FACTOID2025}
\bibfield{author}{\bibinfo{person}{Ningke Li}, \bibinfo{person}{Yahui Song}, \bibinfo{person}{Kailong Wang}, {et~al\mbox{.}}} \bibinfo{year}{2025}\natexlab{b}.
\newblock \showarticletitle{{Detecting LLM Fact-conflicting Hallucinations Enhanced by Temporal-logic-based Reasoning}}.
\newblock \bibinfo{journal}{\emph{ArXiv e-prints}} (\bibinfo{date}{Feb.} \bibinfo{year}{2025}).
\newblock
\href{https://doi.org/10.48550/arXiv.2502.13416}{doi:\nolinkurl{10.48550/arXiv.2502.13416}}


\bibitem[Li et~al\mbox{.}(2024b)]%
        {ReferenceFree2024}
\bibfield{author}{\bibinfo{person}{Qing Li}, \bibinfo{person}{Jiahui Geng}, \bibinfo{person}{Chenyang Lyu}, \bibinfo{person}{Derui Zhu}, \bibinfo{person}{Maxim Panov}, {et~al\mbox{.}}} \bibinfo{year}{2024}\natexlab{b}.
\newblock \showarticletitle{{Reference-free Hallucination Detection for Large Vision-Language Models}}.
\newblock \bibinfo{journal}{\emph{ACL Anthology}} (\bibinfo{date}{Nov.} \bibinfo{year}{2024}), \bibinfo{pages}{4542--4551}.
\newblock
\href{https://doi.org/10.18653/v1/2024.findings-emnlp.262}{doi:\nolinkurl{10.18653/v1/2024.findings-emnlp.262}}


\bibitem[Li et~al\mbox{.}(2025a)]%
        {CounterfactualVLM2025}
\bibfield{author}{\bibinfo{person}{Wei Li}, \bibinfo{person}{Zhen Huang}, \bibinfo{person}{Houqiang Li}, \bibinfo{person}{Le Lu}, \bibinfo{person}{Yang Lu}, \bibinfo{person}{Xinmei Tian}, \bibinfo{person}{Xu Shen}, {and} \bibinfo{person}{Jieping Ye}.} \bibinfo{year}{2025}\natexlab{a}.
\newblock \showarticletitle{Visual Evidence Prompting Mitigates Hallucinations in Large Vision-Language Models}. In \bibinfo{booktitle}{\emph{Proceedings of the 63rd Annual Meeting of the Association for Computational Linguistics (Volume 1: Long Papers)}}, \bibfield{editor}{\bibinfo{person}{Wanxiang Che}, \bibinfo{person}{Joyce Nabende}, \bibinfo{person}{Ekaterina Shutova}, {and} \bibinfo{person}{Mohammad~Taher Pilehvar}} (Eds.). \bibinfo{publisher}{Association for Computational Linguistics}, \bibinfo{address}{Vienna, Austria}, \bibinfo{pages}{4048--4080}.
\newblock
\showISBNx{979-8-89176-251-0}
\href{https://doi.org/10.18653/v1/2025.acl-long.205}{doi:\nolinkurl{10.18653/v1/2025.acl-long.205}}


\bibitem[Li et~al\mbox{.}(2023b)]%
        {POPE2023}
\bibfield{author}{\bibinfo{person}{Yifan Li}, \bibinfo{person}{Yifan Du}, \bibinfo{person}{Kun Zhou}, {et~al\mbox{.}}} \bibinfo{year}{2023}\natexlab{b}.
\newblock \showarticletitle{{Evaluating Object Hallucination in Large Vision-Language Models}}.
\newblock \bibinfo{journal}{\emph{ArXiv e-prints}} (\bibinfo{date}{May} \bibinfo{year}{2023}).
\newblock
\href{https://doi.org/10.48550/arXiv.2305.10355}{doi:\nolinkurl{10.48550/arXiv.2305.10355}}


\bibitem[Liao et~al\mbox{.}(2025)]%
        {VASE2025}
\bibfield{author}{\bibinfo{person}{Zehui Liao}, \bibinfo{person}{Shishuai Hu}, \bibinfo{person}{Ke Zou}, \bibinfo{person}{Huazhu Fu}, \bibinfo{person}{Liangli Zhen}, {et~al\mbox{.}}} \bibinfo{year}{2025}\natexlab{}.
\newblock \showarticletitle{{Vision-Amplified Semantic Entropy for Hallucination Detection in Medical Visual Question Answering}}.
\newblock \bibinfo{journal}{\emph{ArXiv e-prints}} (\bibinfo{date}{March} \bibinfo{year}{2025}).
\newblock
\href{https://doi.org/10.48550/arXiv.2503.20504}{doi:\nolinkurl{10.48550/arXiv.2503.20504}}


\bibitem[Liu et~al\mbox{.}(2022)]%
        {HaDes2023}
\bibfield{author}{\bibinfo{person}{Tianyu Liu}, \bibinfo{person}{Yizhe Zhang}, \bibinfo{person}{Chris Brockett}, \bibinfo{person}{Yi Mao}, \bibinfo{person}{Zhifang Sui}, \bibinfo{person}{Weizhu Chen}, {and} \bibinfo{person}{Bill Dolan}.} \bibinfo{year}{2022}\natexlab{}.
\newblock \showarticletitle{A Token-level Reference-free Hallucination Detection Benchmark for Free-form Text Generation}. In \bibinfo{booktitle}{\emph{Proceedings of the 60th Annual Meeting of the Association for Computational Linguistics (Volume 1: Long Papers)}}, \bibfield{editor}{\bibinfo{person}{Smaranda Muresan}, \bibinfo{person}{Preslav Nakov}, {and} \bibinfo{person}{Aline Villavicencio}} (Eds.). \bibinfo{publisher}{Association for Computational Linguistics}, \bibinfo{address}{Dublin, Ireland}, \bibinfo{pages}{6723--6737}.
\newblock
\href{https://doi.org/10.18653/v1/2022.acl-long.464}{doi:\nolinkurl{10.18653/v1/2022.acl-long.464}}


\bibitem[Liu et~al\mbox{.}(2023)]%
        {Liu2023GEval}
\bibfield{author}{\bibinfo{person}{Yang Liu}, \bibinfo{person}{Dan Iter}, \bibinfo{person}{Yichong Xu}, \bibinfo{person}{Shuohang Wang}, \bibinfo{person}{Ruochen Xu}, {and} \bibinfo{person}{Chenguang Zhu}.} \bibinfo{year}{2023}\natexlab{}.
\newblock \showarticletitle{{G}-Eval: {NLG} Evaluation using Gpt-4 with Better Human Alignment}. In \bibinfo{booktitle}{\emph{Proceedings of the 2023 Conference on Empirical Methods in Natural Language Processing}}, \bibfield{editor}{\bibinfo{person}{Houda Bouamor}, \bibinfo{person}{Juan Pino}, {and} \bibinfo{person}{Kalika Bali}} (Eds.). \bibinfo{publisher}{Association for Computational Linguistics}, \bibinfo{address}{Singapore}, \bibinfo{pages}{2511--2522}.
\newblock
\href{https://doi.org/10.18653/v1/2023.emnlp-main.153}{doi:\nolinkurl{10.18653/v1/2023.emnlp-main.153}}


\bibitem[Luo et~al\mbox{.}(2024)]%
        {HalluDial2024}
\bibfield{author}{\bibinfo{person}{Wen Luo}, \bibinfo{person}{Tianshu Shen}, \bibinfo{person}{Wei Li}, {et~al\mbox{.}}} \bibinfo{year}{2024}\natexlab{}.
\newblock \showarticletitle{{HalluDial: A Large-Scale Benchmark for Automatic Dialogue-Level Hallucination Evaluation}}.
\newblock \bibinfo{journal}{\emph{ArXiv e-prints}} (\bibinfo{date}{June} \bibinfo{year}{2024}).
\newblock
\href{https://doi.org/10.48550/arXiv.2406.07070}{doi:\nolinkurl{10.48550/arXiv.2406.07070}}


\bibitem[Ma et~al\mbox{.}(2018)]%
        {ma2018awe}
\bibfield{author}{\bibinfo{person}{Tengfei Ma}, \bibinfo{person}{Chiamin Wu}, \bibinfo{person}{Cao Xiao}, {et~al\mbox{.}}} \bibinfo{year}{2018}\natexlab{}.
\newblock \showarticletitle{{AWE: Asymmetric Word Embedding for Textual Entailment}}.
\newblock \bibinfo{journal}{\emph{ArXiv e-prints}} (\bibinfo{date}{Sept.} \bibinfo{year}{2018}).
\newblock
\href{https://doi.org/10.48550/arXiv.1809.04047}{doi:\nolinkurl{10.48550/arXiv.1809.04047}}


\bibitem[Mallen et~al\mbox{.}(2023)]%
        {PopQA2023}
\bibfield{author}{\bibinfo{person}{Alex Mallen}, \bibinfo{person}{Akari Asai}, \bibinfo{person}{Victor Zhong}, \bibinfo{person}{Rajarshi Das}, \bibinfo{person}{Daniel Khashabi}, {and} \bibinfo{person}{Hannaneh Hajishirzi}.} \bibinfo{year}{2023}\natexlab{}.
\newblock \showarticletitle{When Not to Trust Language Models: Investigating Effectiveness of Parametric and Non-Parametric Memories}. In \bibinfo{booktitle}{\emph{Proceedings of the 61st Annual Meeting of the Association for Computational Linguistics (Volume 1: Long Papers)}}, \bibfield{editor}{\bibinfo{person}{Anna Rogers}, \bibinfo{person}{Jordan Boyd-Graber}, {and} \bibinfo{person}{Naoaki Okazaki}} (Eds.). \bibinfo{publisher}{Association for Computational Linguistics}, \bibinfo{address}{Toronto, Canada}, \bibinfo{pages}{9802--9822}.
\newblock
\href{https://doi.org/10.18653/v1/2023.acl-long.546}{doi:\nolinkurl{10.18653/v1/2023.acl-long.546}}


\bibitem[Manakul et~al\mbox{.}(2023)]%
        {Manakul2023SelfCheckGPT}
\bibfield{author}{\bibinfo{person}{Potsawee Manakul}, \bibinfo{person}{Adian Liusie}, {and} \bibinfo{person}{Mark Gales}.} \bibinfo{year}{2023}\natexlab{}.
\newblock \showarticletitle{{S}elf{C}heck{GPT}: Zero-Resource Black-Box Hallucination Detection for Generative Large Language Models}. In \bibinfo{booktitle}{\emph{Proceedings of the 2023 Conference on Empirical Methods in Natural Language Processing}}, \bibfield{editor}{\bibinfo{person}{Houda Bouamor}, \bibinfo{person}{Juan Pino}, {and} \bibinfo{person}{Kalika Bali}} (Eds.). \bibinfo{publisher}{Association for Computational Linguistics}, \bibinfo{address}{Singapore}, \bibinfo{pages}{9004--9017}.
\newblock
\href{https://doi.org/10.18653/v1/2023.emnlp-main.557}{doi:\nolinkurl{10.18653/v1/2023.emnlp-main.557}}


\bibitem[Mikolov et~al\mbox{.}(2013)]%
        {mikolov2013efficient}
\bibfield{author}{\bibinfo{person}{Tomas Mikolov}, \bibinfo{person}{Kai Chen}, \bibinfo{person}{Greg Corrado}, {and} \bibinfo{person}{Jeffrey Dean}.} \bibinfo{year}{2013}\natexlab{}.
\newblock \showarticletitle{Efficient Estimation of Word Representations in Vector Space}. In \bibinfo{booktitle}{\emph{Proceedings of the International Conference on Learning Representations (ICLR)}}.
\newblock
\urldef\tempurl%
\url{https://arxiv.org/abs/1301.3781}
\showURL{%
\tempurl}


\bibitem[Minderer et~al\mbox{.}(2022)]%
        {DetectorVoting2025}
\bibfield{author}{\bibinfo{person}{Matthias Minderer}, \bibinfo{person}{Alexey Gritsenko}, \bibinfo{person}{Austin Stone}, {et~al\mbox{.}}} \bibinfo{year}{2022}\natexlab{}.
\newblock \showarticletitle{{Simple Open-Vocabulary Object Detection}}.
\newblock In \bibinfo{booktitle}{\emph{{Computer Vision {\textendash} ECCV 2022}}}. \bibinfo{publisher}{Springer}, \bibinfo{address}{Cham, Switzerland}, \bibinfo{pages}{728--755}.
\newblock
\showISBNx{978-3-031-20080-9}
\href{https://doi.org/10.1007/978-3-031-20080-9_42}{doi:\nolinkurl{10.1007/978-3-031-20080-9_42}}


\bibitem[Padhi et~al\mbox{.}(2025)]%
        {padhi2025multimodal}
\bibfield{author}{\bibinfo{person}{Trilok Padhi}, \bibinfo{person}{Ramneet Kaur}, \bibinfo{person}{Adam~D. Cobb}, {et~al\mbox{.}}} \bibinfo{year}{2025}\natexlab{}.
\newblock \showarticletitle{{Calibrating Uncertainty Quantification of Multi-Modal LLMs using Grounding}}.
\newblock \bibinfo{journal}{\emph{ArXiv e-prints}} (\bibinfo{date}{April} \bibinfo{year}{2025}).
\newblock
\href{https://doi.org/10.48550/arXiv.2505.03788}{doi:\nolinkurl{10.48550/arXiv.2505.03788}}


\bibitem[Pagnoni et~al\mbox{.}(2021)]%
        {FRANK2021}
\bibfield{author}{\bibinfo{person}{Artidoro Pagnoni}, \bibinfo{person}{Vidhisha Balachandran}, {and} \bibinfo{person}{Yulia Tsvetkov}.} \bibinfo{year}{2021}\natexlab{}.
\newblock \showarticletitle{{Understanding Factuality in Abstractive Summarization with FRANK: A Benchmark for Factuality Metrics}}.
\newblock \bibinfo{journal}{\emph{ArXiv e-prints}} (\bibinfo{date}{April} \bibinfo{year}{2021}).
\newblock
\href{https://doi.org/10.48550/arXiv.2104.13346}{doi:\nolinkurl{10.48550/arXiv.2104.13346}}


\bibitem[Pal et~al\mbox{.}(2023)]%
        {MedHal2023}
\bibfield{author}{\bibinfo{person}{Ankit Pal}, \bibinfo{person}{Logesh~Kumar Umapathi}, {and} \bibinfo{person}{Malaikannan Sankarasubbu}.} \bibinfo{year}{2023}\natexlab{}.
\newblock \showarticletitle{{Med-HALT: Medical Domain Hallucination Test for Large Language Models}}.
\newblock \bibinfo{journal}{\emph{ACL Anthology}} (\bibinfo{date}{Dec.} \bibinfo{year}{2023}), \bibinfo{pages}{314--334}.
\newblock
\href{https://doi.org/10.18653/v1/2023.conll-1.21}{doi:\nolinkurl{10.18653/v1/2023.conll-1.21}}


\bibitem[Petryk et~al\mbox{.}(2024)]%
        {ALOHa2024}
\bibfield{author}{\bibinfo{person}{Suzanne Petryk}, \bibinfo{person}{David~M. Chan}, \bibinfo{person}{Anish Kachinthaya}, \bibinfo{person}{Haodi Zou}, \bibinfo{person}{John Canny}, \bibinfo{person}{Joseph~E. Gonzalez}, {and} \bibinfo{person}{Trevor Darrell}.} \bibinfo{year}{2024}\natexlab{}.
\newblock \showarticletitle{{ALOH}a: A New Measure for Hallucination in Captioning Models}. In \bibinfo{booktitle}{\emph{Proceedings of the 2024 Conference of the North American Chapter of the Association for Computational Linguistics: Human Language Technologies (Volume 2: Short Papers)}}, \bibfield{editor}{\bibinfo{person}{Kevin Duh}, \bibinfo{person}{Helena Gomez}, {and} \bibinfo{person}{Steven Bethard}} (Eds.). \bibinfo{publisher}{Association for Computational Linguistics}, \bibinfo{address}{Mexico City, Mexico}, \bibinfo{pages}{342--357}.
\newblock
\href{https://doi.org/10.18653/v1/2024.naacl-short.30}{doi:\nolinkurl{10.18653/v1/2024.naacl-short.30}}


\bibitem[Pham and Schott(2024)]%
        {HPOPE2024}
\bibfield{author}{\bibinfo{person}{Nhi Pham} {and} \bibinfo{person}{Michael Schott}.} \bibinfo{year}{2024}\natexlab{}.
\newblock \showarticletitle{{H-POPE: Hierarchical Polling-based Probing Evaluation of Hallucinations in Large Vision-Language Models}}.
\newblock \bibinfo{journal}{\emph{ArXiv e-prints}} (\bibinfo{date}{Nov.} \bibinfo{year}{2024}).
\newblock
\href{https://doi.org/10.48550/arXiv.2411.04077}{doi:\nolinkurl{10.48550/arXiv.2411.04077}}


\bibitem[Prabhakaran et~al\mbox{.}(2025)]%
        {VADE2025}
\bibfield{author}{\bibinfo{person}{Vishnu Prabhakaran}, \bibinfo{person}{Purav Aggarwal}, \bibinfo{person}{Vinay~Kumar Verma}, \bibinfo{person}{Gokul Swamy}, {and} \bibinfo{person}{Anoop Saladi}.} \bibinfo{year}{2025}\natexlab{}.
\newblock \showarticletitle{{VADE}: Visual Attention Guided Hallucination Detection and Elimination}. In \bibinfo{booktitle}{\emph{Findings of the Association for Computational Linguistics: ACL 2025}}, \bibfield{editor}{\bibinfo{person}{Wanxiang Che}, \bibinfo{person}{Joyce Nabende}, \bibinfo{person}{Ekaterina Shutova}, {and} \bibinfo{person}{Mohammad~Taher Pilehvar}} (Eds.). \bibinfo{publisher}{Association for Computational Linguistics}, \bibinfo{address}{Vienna, Austria}, \bibinfo{pages}{14949--14965}.
\newblock
\showISBNx{979-8-89176-256-5}
\href{https://doi.org/10.18653/v1/2025.findings-acl.773}{doi:\nolinkurl{10.18653/v1/2025.findings-acl.773}}


\bibitem[Qi et~al\mbox{.}(2024)]%
        {AHESurvey2024}
\bibfield{author}{\bibinfo{person}{Siya Qi}, \bibinfo{person}{Lin Gui}, \bibinfo{person}{Yulan He}, {et~al\mbox{.}}} \bibinfo{year}{2024}\natexlab{}.
\newblock \showarticletitle{{A Survey of Automatic Hallucination Evaluation on Natural Language Generation}}.
\newblock \bibinfo{journal}{\emph{ArXiv e-prints}} (\bibinfo{date}{April} \bibinfo{year}{2024}).
\newblock
\href{https://doi.org/10.48550/arXiv.2404.12041}{doi:\nolinkurl{10.48550/arXiv.2404.12041}}


\bibitem[Radford et~al\mbox{.}(2021)]%
        {radford2021learning}
\bibfield{author}{\bibinfo{person}{Alec Radford}, \bibinfo{person}{Jong~Wook Kim}, \bibinfo{person}{Chris Hallacy}, \bibinfo{person}{Aditya Ramesh}, \bibinfo{person}{Gabriel Goh}, \bibinfo{person}{Sandhini Agarwal}, \bibinfo{person}{Girish Sastry}, \bibinfo{person}{Amanda Askell}, \bibinfo{person}{Pamela Mishkin}, \bibinfo{person}{Jack Clark}, \bibinfo{person}{Gretchen Krueger}, {and} \bibinfo{person}{Ilya Sutskever}.} \bibinfo{year}{2021}\natexlab{}.
\newblock \showarticletitle{Learning Transferable Visual Models from Natural Language Supervision}. In \bibinfo{booktitle}{\emph{Proceedings of the 38th International Conference on Machine Learning (ICML)}}. \bibinfo{publisher}{PMLR}, \bibinfo{pages}{8748--8763}.
\newblock
\urldef\tempurl%
\url{https://proceedings.mlr.press/v139/radford21a.html}
\showURL{%
\tempurl}


\bibitem[Reimers and Gurevych(2019)]%
        {reimers2019sentence}
\bibfield{author}{\bibinfo{person}{Nils Reimers} {and} \bibinfo{person}{Iryna Gurevych}.} \bibinfo{year}{2019}\natexlab{}.
\newblock \showarticletitle{Sentence-BERT: Sentence Embeddings using Siamese BERT Networks}. In \bibinfo{booktitle}{\emph{Proceedings of the 2019 Conference on Empirical Methods in Natural Language Processing and the 9th International Joint Conference on Natural Language Processing (EMNLP–IJCNLP)}}. Association for Computational Linguistics, \bibinfo{address}{Hong Kong, China}, \bibinfo{pages}{3982--3992}.
\newblock
\href{https://doi.org/10.18653/v1/D19-1410}{doi:\nolinkurl{10.18653/v1/D19-1410}}


\bibitem[Renze(2024)]%
        {TemperatureamplingRenze2024}
\bibfield{author}{\bibinfo{person}{Matthew Renze}.} \bibinfo{year}{2024}\natexlab{}.
\newblock \showarticletitle{{The Effect of Sampling Temperature on Problem Solving in Large Language Models}}.
\newblock \bibinfo{journal}{\emph{ACL Anthology}} (\bibinfo{date}{Nov.} \bibinfo{year}{2024}), \bibinfo{pages}{7346--7356}.
\newblock
\href{https://doi.org/10.18653/v1/2024.findings-emnlp.432}{doi:\nolinkurl{10.18653/v1/2024.findings-emnlp.432}}


\bibitem[Rohrbach et~al\mbox{.}(2018)]%
        {ObjectHalBench2024}
\bibfield{author}{\bibinfo{person}{Anna Rohrbach}, \bibinfo{person}{Lisa~Anne Hendricks}, \bibinfo{person}{Kaylee Burns}, {et~al\mbox{.}}} \bibinfo{year}{2018}\natexlab{}.
\newblock \showarticletitle{{Object Hallucination in Image Captioning}}.
\newblock \bibinfo{journal}{\emph{ACL Anthology}} (\bibinfo{year}{2018}), \bibinfo{pages}{4035--4045}.
\newblock
\href{https://doi.org/10.18653/v1/D18-1437}{doi:\nolinkurl{10.18653/v1/D18-1437}}


\bibitem[Romanov and Shivade(2018)]%
        {Romanov2018}
\bibfield{author}{\bibinfo{person}{Alexey Romanov} {and} \bibinfo{person}{Chaitanya Shivade}.} \bibinfo{year}{2018}\natexlab{}.
\newblock \showarticletitle{{Lessons from Natural Language Inference in the Clinical Domain}}.
\newblock \bibinfo{journal}{\emph{ACL Anthology}} (\bibinfo{year}{2018}), \bibinfo{pages}{1586--1596}.
\newblock
\href{https://doi.org/10.18653/v1/D18-1187}{doi:\nolinkurl{10.18653/v1/D18-1187}}


\bibitem[Sardana(2025)]%
        {RAGSurvey2025}
\bibfield{author}{\bibinfo{person}{Ashish Sardana}.} \bibinfo{year}{2025}\natexlab{}.
\newblock \showarticletitle{{Real-Time Evaluation Models for RAG: Who Detects Hallucinations Best?}}
\newblock \bibinfo{journal}{\emph{ArXiv e-prints}} (\bibinfo{date}{March} \bibinfo{year}{2025}).
\newblock
\href{https://doi.org/10.48550/arXiv.2503.21157}{doi:\nolinkurl{10.48550/arXiv.2503.21157}}


\bibitem[Sellergren et~al\mbox{.}(2025)]%
        {MedGemma}
\bibfield{author}{\bibinfo{person}{Andrew Sellergren}, \bibinfo{person}{Sahar Kazemzadeh}, \bibinfo{person}{Tiam Jaroensri}, {et~al\mbox{.}}} \bibinfo{year}{2025}\natexlab{}.
\newblock \showarticletitle{{MedGemma Technical Report}}.
\newblock \bibinfo{journal}{\emph{ArXiv e-prints}} (\bibinfo{date}{July} \bibinfo{year}{2025}).
\newblock
\href{https://doi.org/10.48550/arXiv.2507.05201}{doi:\nolinkurl{10.48550/arXiv.2507.05201}}


\bibitem[Shannon(1948)]%
        {Shannon1948}
\bibfield{author}{\bibinfo{person}{Claude~E Shannon}.} \bibinfo{year}{1948}\natexlab{}.
\newblock \showarticletitle{A mathematical theory of communication}.
\newblock \bibinfo{journal}{\emph{The Bell system technical journal}} \bibinfo{volume}{27}, \bibinfo{number}{3} (\bibinfo{year}{1948}), \bibinfo{pages}{379--423}.
\newblock


\bibitem[Tu et~al\mbox{.}(2024)]%
        {ODE2024}
\bibfield{author}{\bibinfo{person}{Yahan Tu}, \bibinfo{person}{Rui Hu}, {and} \bibinfo{person}{Jitao Sang}.} \bibinfo{year}{2024}\natexlab{}.
\newblock \showarticletitle{{ODE: Open-Set Evaluation of Hallucinations in Multimodal Large Language Models}}.
\newblock \bibinfo{journal}{\emph{ArXiv e-prints}} (\bibinfo{date}{Sept.} \bibinfo{year}{2024}).
\newblock
\href{https://doi.org/10.48550/arXiv.2409.09318}{doi:\nolinkurl{10.48550/arXiv.2409.09318}}


\bibitem[Van~der Maaten and Hinton(2008)]%
        {vandermaaten2008visualizing}
\bibfield{author}{\bibinfo{person}{Laurens Van~der Maaten} {and} \bibinfo{person}{Geoffrey Hinton}.} \bibinfo{year}{2008}\natexlab{}.
\newblock \showarticletitle{Visualizing Data using t-SNE}.
\newblock \bibinfo{journal}{\emph{Journal of Machine Learning Research}} \bibinfo{volume}{9}, \bibinfo{number}{86} (\bibinfo{year}{2008}), \bibinfo{pages}{2579--2605}.
\newblock
\urldef\tempurl%
\url{https://www.jmlr.org/papers/v9/vandermaaten08a.html}
\showURL{%
\tempurl}


\bibitem[Vashurin et~al\mbox{.}(2025)]%
        {vashurin2025benchmarking}
\bibfield{author}{\bibinfo{person}{Roman Vashurin}, \bibinfo{person}{Ekaterina Fadeeva}, \bibinfo{person}{Artem Vazhentsev}, \bibinfo{person}{Lyudmila Rvanova}, \bibinfo{person}{Daniil Vasilev}, {et~al\mbox{.}}} \bibinfo{year}{2025}\natexlab{}.
\newblock \showarticletitle{{Benchmarking Uncertainty Quantification Methods for Large Language Models with LM-Polygraph}}.
\newblock \bibinfo{journal}{\emph{Transactions of the Association for Computational Linguistics}}  \bibinfo{volume}{13} (\bibinfo{date}{Sept.} \bibinfo{year}{2025}), \bibinfo{pages}{220--248}.
\newblock
\href{https://doi.org/10.1162/tacl_a_00737}{doi:\nolinkurl{10.1162/tacl_a_00737}}


\bibitem[Vu et~al\mbox{.}(2023)]%
        {FreshQA2024}
\bibfield{author}{\bibinfo{person}{Tu Vu}, \bibinfo{person}{Mohit Iyyer}, \bibinfo{person}{Xuezhi Wang}, {et~al\mbox{.}}} \bibinfo{year}{2023}\natexlab{}.
\newblock \showarticletitle{{FreshLLMs: Refreshing Large Language Models with Search Engine Augmentation}}.
\newblock \bibinfo{journal}{\emph{ArXiv e-prints}} (\bibinfo{date}{Oct.} \bibinfo{year}{2023}).
\newblock
\href{https://doi.org/10.48550/arXiv.2310.03214}{doi:\nolinkurl{10.48550/arXiv.2310.03214}}


\bibitem[Wada et~al\mbox{.}(2024)]%
        {Polos2024}
\bibfield{author}{\bibinfo{person}{Yuiga Wada}, \bibinfo{person}{Kanta Kaneda}, \bibinfo{person}{Daichi Saito}, {et~al\mbox{.}}} \bibinfo{year}{2024}\natexlab{}.
\newblock \showarticletitle{{Polos: Multimodal Metric Learning from Human Feedback for Image Captioning}}.
\newblock \bibinfo{journal}{\emph{ArXiv e-prints}} (\bibinfo{date}{Feb.} \bibinfo{year}{2024}).
\newblock
\href{https://doi.org/10.48550/arXiv.2402.18091}{doi:\nolinkurl{10.48550/arXiv.2402.18091}}


\bibitem[Wang and Ji(2024)]%
        {epistemic-uq2024}
\bibfield{author}{\bibinfo{person}{Hanjing Wang} {and} \bibinfo{person}{Qiang Ji}.} \bibinfo{year}{2024}\natexlab{}.
\newblock \showarticletitle{{Epistemic Uncertainty Quantification For Pre-trained Neural Network}}.
\newblock \bibinfo{journal}{\emph{ArXiv e-prints}} (\bibinfo{date}{April} \bibinfo{year}{2024}).
\newblock
\href{https://doi.org/10.48550/arXiv.2404.10124}{doi:\nolinkurl{10.48550/arXiv.2404.10124}}


\bibitem[Wang et~al\mbox{.}(2023)]%
        {AMBER2024}
\bibfield{author}{\bibinfo{person}{Junyang Wang}, \bibinfo{person}{Yuhang Wang}, \bibinfo{person}{Guohai Xu}, {et~al\mbox{.}}} \bibinfo{year}{2023}\natexlab{}.
\newblock \showarticletitle{{AMBER: An LLM-free Multi-dimensional Benchmark for MLLMs Hallucination Evaluation}}.
\newblock \bibinfo{journal}{\emph{ArXiv e-prints}} (\bibinfo{date}{Nov.} \bibinfo{year}{2023}).
\newblock
\href{https://doi.org/10.48550/arXiv.2311.07397}{doi:\nolinkurl{10.48550/arXiv.2311.07397}}


\bibitem[Wei et~al\mbox{.}(2024a)]%
        {SimpleQA2022}
\bibfield{author}{\bibinfo{person}{Jason Wei}, \bibinfo{person}{Nguyen Karina}, \bibinfo{person}{Hyung~Won Chung}, {et~al\mbox{.}}} \bibinfo{year}{2024}\natexlab{a}.
\newblock \showarticletitle{{Measuring short-form factuality in large language models}}.
\newblock \bibinfo{journal}{\emph{ArXiv e-prints}} (\bibinfo{date}{Nov.} \bibinfo{year}{2024}).
\newblock
\href{https://doi.org/10.48550/arXiv.2411.04368}{doi:\nolinkurl{10.48550/arXiv.2411.04368}}


\bibitem[Wei et~al\mbox{.}(2024b)]%
        {LongFact2024}
\bibfield{author}{\bibinfo{person}{Jerry Wei}, \bibinfo{person}{Chengrun Yang}, \bibinfo{person}{Xinying Song}, {et~al\mbox{.}}} \bibinfo{year}{2024}\natexlab{b}.
\newblock \showarticletitle{{Long-form factuality in large language models}}.
\newblock \bibinfo{journal}{\emph{ArXiv e-prints}} (\bibinfo{date}{March} \bibinfo{year}{2024}).
\newblock
\href{https://doi.org/10.48550/arXiv.2403.18802}{doi:\nolinkurl{10.48550/arXiv.2403.18802}}


\bibitem[Wienholt et~al\mbox{.}(2025)]%
        {DSE2025}
\bibfield{author}{\bibinfo{person}{Patrick Wienholt}, \bibinfo{person}{Sophie Caselitz}, \bibinfo{person}{Robert Siepmann}, {et~al\mbox{.}}} \bibinfo{year}{2025}\natexlab{}.
\newblock \showarticletitle{{Hallucination Filtering in Radiology Vision-Language Models Using Discrete Semantic Entropy}}.
\newblock \bibinfo{journal}{\emph{ArXiv e-prints}} (\bibinfo{date}{Oct.} \bibinfo{year}{2025}).
\newblock
\href{https://doi.org/10.48550/arXiv.2510.09256}{doi:\nolinkurl{10.48550/arXiv.2510.09256}}


\bibitem[Williams et~al\mbox{.}(2018)]%
        {MNLI2018}
\bibfield{author}{\bibinfo{person}{Adina Williams}, \bibinfo{person}{Nikita Nangia}, {and} \bibinfo{person}{Samuel Bowman}.} \bibinfo{year}{2018}\natexlab{}.
\newblock \showarticletitle{{A Broad-Coverage Challenge Corpus for Sentence Understanding through Inference}}.
\newblock \bibinfo{journal}{\emph{ACL Anthology}} (\bibinfo{date}{June} \bibinfo{year}{2018}), \bibinfo{pages}{1112--1122}.
\newblock
\href{https://doi.org/10.18653/v1/N18-1101}{doi:\nolinkurl{10.18653/v1/N18-1101}}


\bibitem[Wu et~al\mbox{.}(2024)]%
        {HALTMedVQA2024}
\bibfield{author}{\bibinfo{person}{Jinge Wu}, \bibinfo{person}{Yunsoo Kim}, {and} \bibinfo{person}{Honghan Wu}.} \bibinfo{year}{2024}\natexlab{}.
\newblock \showarticletitle{{Hallucination Benchmark in Medical Visual Question Answering}}.
\newblock \bibinfo{journal}{\emph{ArXiv e-prints}} (\bibinfo{date}{Jan.} \bibinfo{year}{2024}).
\newblock
\href{https://doi.org/10.48550/arXiv.2401.05827}{doi:\nolinkurl{10.48550/arXiv.2401.05827}}


\bibitem[Yang et~al\mbox{.}(2025a)]%
        {Qwen3Yang2025May}
\bibfield{author}{\bibinfo{person}{An Yang}, \bibinfo{person}{Anfeng Li}, \bibinfo{person}{Baosong Yang}, {et~al\mbox{.}}} \bibinfo{year}{2025}\natexlab{a}.
\newblock \showarticletitle{{Qwen3 Technical Report}}.
\newblock \bibinfo{journal}{\emph{ArXiv e-prints}} (\bibinfo{date}{May} \bibinfo{year}{2025}).
\newblock
\href{https://doi.org/10.48550/arXiv.2505.09388}{doi:\nolinkurl{10.48550/arXiv.2505.09388}}


\bibitem[Yang et~al\mbox{.}(2025b)]%
        {MetaQA2025}
\bibfield{author}{\bibinfo{person}{Borui Yang}, \bibinfo{person}{Md~Afif~Al Mamun}, \bibinfo{person}{Jie~M. Zhang}, {et~al\mbox{.}}} \bibinfo{year}{2025}\natexlab{b}.
\newblock \showarticletitle{{Hallucination Detection in Large Language Models with Metamorphic Relations}}.
\newblock \bibinfo{journal}{\emph{ArXiv e-prints}} (\bibinfo{date}{Feb.} \bibinfo{year}{2025}).
\newblock
\href{https://doi.org/10.48550/arXiv.2502.15844}{doi:\nolinkurl{10.48550/arXiv.2502.15844}}


\bibitem[Yeh et~al\mbox{.}(2025)]%
        {HalluEntity2024}
\bibfield{author}{\bibinfo{person}{Min-Hsuan Yeh}, \bibinfo{person}{Max Kamachee}, \bibinfo{person}{Seongheon Park}, {et~al\mbox{.}}} \bibinfo{year}{2025}\natexlab{}.
\newblock \showarticletitle{{HalluEntity: Benchmarking and Understanding Entity-Level Hallucination Detection}}.
\newblock \bibinfo{journal}{\emph{ArXiv e-prints}} (\bibinfo{date}{Feb.} \bibinfo{year}{2025}).
\newblock
\href{https://doi.org/10.48550/arXiv.2502.11948}{doi:\nolinkurl{10.48550/arXiv.2502.11948}}


\bibitem[Yu et~al\mbox{.}(2023)]%
        {HalluciDoctor2023}
\bibfield{author}{\bibinfo{person}{Qifan Yu}, \bibinfo{person}{Juncheng Li}, \bibinfo{person}{Longhui Wei}, \bibinfo{person}{Liang Pang}, \bibinfo{person}{Wentao Ye}, {et~al\mbox{.}}} \bibinfo{year}{2023}\natexlab{}.
\newblock \showarticletitle{{HalluciDoctor: Mitigating Hallucinatory Toxicity in Visual Instruction Data}}.
\newblock \bibinfo{journal}{\emph{ArXiv e-prints}} (\bibinfo{date}{Nov.} \bibinfo{year}{2023}).
\newblock
\href{https://doi.org/10.48550/arXiv.2311.13614}{doi:\nolinkurl{10.48550/arXiv.2311.13614}}


\bibitem[Zha et~al\mbox{.}(2023)]%
        {DialFact2022}
\bibfield{author}{\bibinfo{person}{Yuheng Zha}, \bibinfo{person}{Yichi Yang}, \bibinfo{person}{Ruichen Li}, {and} \bibinfo{person}{Zhiting Hu}.} \bibinfo{year}{2023}\natexlab{}.
\newblock \showarticletitle{{A}lign{S}core: Evaluating Factual Consistency with A Unified Alignment Function}. In \bibinfo{booktitle}{\emph{Proceedings of the 61st Annual Meeting of the Association for Computational Linguistics (Volume 1: Long Papers)}}, \bibfield{editor}{\bibinfo{person}{Anna Rogers}, \bibinfo{person}{Jordan Boyd-Graber}, {and} \bibinfo{person}{Naoaki Okazaki}} (Eds.). \bibinfo{publisher}{Association for Computational Linguistics}, \bibinfo{address}{Toronto, Canada}, \bibinfo{pages}{11328--11348}.
\newblock
\href{https://doi.org/10.18653/v1/2023.acl-long.634}{doi:\nolinkurl{10.18653/v1/2023.acl-long.634}}


\bibitem[Zhang et~al\mbox{.}(2024)]%
        {RadFlag2024}
\bibfield{author}{\bibinfo{person}{Serena Zhang}, \bibinfo{person}{Sraavya Sambara}, \bibinfo{person}{Oishi Banerjee}, \bibinfo{person}{Julian Acosta}, \bibinfo{person}{L.~John Fahrner}, {et~al\mbox{.}}} \bibinfo{year}{2024}\natexlab{}.
\newblock \showarticletitle{{RadFlag: A Black-Box Hallucination Detection Method for Medical Vision Language Models}}.
\newblock \bibinfo{journal}{\emph{ArXiv e-prints}} (\bibinfo{date}{Nov.} \bibinfo{year}{2024}).
\newblock
\href{https://doi.org/10.48550/arXiv.2411.00299}{doi:\nolinkurl{10.48550/arXiv.2411.00299}}


\bibitem[Zhu et~al\mbox{.}(2025)]%
        {AttentionCalibration2025}
\bibfield{author}{\bibinfo{person}{Younan Zhu}, \bibinfo{person}{Linwei Tao}, \bibinfo{person}{Minjing Dong}, {et~al\mbox{.}}} \bibinfo{year}{2025}\natexlab{}.
\newblock \showarticletitle{{Mitigating Object Hallucinations in Large Vision-Language Models via Attention Calibration}}.
\newblock \bibinfo{journal}{\emph{ArXiv e-prints}} (\bibinfo{date}{Feb.} \bibinfo{year}{2025}).
\newblock
\href{https://doi.org/10.48550/arXiv.2502.01969}{doi:\nolinkurl{10.48550/arXiv.2502.01969}}


\bibitem[Zhu et~al\mbox{.}(2024)]%
        {WildHallucinations2024}
\bibfield{author}{\bibinfo{person}{Zhiying Zhu}, \bibinfo{person}{Yiming Yang}, {and} \bibinfo{person}{Zhiqing Sun}.} \bibinfo{year}{2024}\natexlab{}.
\newblock \showarticletitle{{HaluEval-Wild: Evaluating Hallucinations of Language Models in the Wild}}.
\newblock \bibinfo{journal}{\emph{ArXiv e-prints}} (\bibinfo{date}{March} \bibinfo{year}{2024}).
\newblock
\href{https://doi.org/10.48550/arXiv.2403.04307}{doi:\nolinkurl{10.48550/arXiv.2403.04307}}


\end{thebibliography}

\clearpage
\onecolumn
\appendix
\section*{Appendix}

\section{Comparison of NLI and Embedding Clustering Results}

We present two \textit{extreme} examples demonstrating cases where the Natural Language Inference (NLI)–based clustering and the Embedding-based clustering methods produce substantially different groupings. In each example, the complete cluster contents are shown. The symbol $\rightarrow$ indicates the beginning of an individual sample response generated by the vision–language model (VLM). Colors denote the source of each response: \textcolor{blue}{blue} corresponds to outputs from the \textit{low-temperature (original image)} setting, \textcolor{cyan}{cyan} represents \textit{high-temperature responses from the original image}, and \textcolor{orange}{orange} indicates \textit{high-temperature responses generated from distorted-image inputs}.

\subsection{Example 1: "No abnormality" statements}

\noindent
\begin{tabular}{@{}m{0.72\textwidth} m{0.26\textwidth}@{}}
\raggedright
\textit{VQA Source:} VQA-RAD Test Set\\
\textit{Question:} Is there any intraparenchymal abnormalities in the lung fields?\\
\textit{Reference answer:} No\\
\textit{Answer variant requested:} Concise\\
\textit{Model:} Qwen2.5-VL-7B-Instruct\\
&
\centering
\includegraphics[width=0.5\linewidth]{figures/images/radvqa_ShortClinical_453.png}
\end{tabular}\\

\textbf{NLI clustering}\\
C-0: \textcolor{blue}{$\rightarrow$~No} \textcolor{cyan}{$\rightarrow$~no} \textcolor{cyan}{$\rightarrow$~No evident abnormalities} \textcolor{cyan}{$\rightarrow$~No, no evident focal opacities} \textcolor{cyan}{$\rightarrow$~no} \textcolor{cyan}{$\rightarrow$~No} \textcolor{cyan}{$\rightarrow$~No} \textcolor{orange}{$\rightarrow$~no} \textcolor{orange}{$\rightarrow$~no}\\
C-1: \textcolor{cyan}{$\rightarrow$~No visible abnormalities} \textcolor{cyan}{$\rightarrow$~No evidenced abnormality} \textcolor{cyan}{$\rightarrow$~No palpable evidence found} \textcolor{cyan}{$\rightarrow$~No admireisible intraparenchymal abnormalities found.} \textcolor{orange}{$\rightarrow$~No obvious abnormalities noted} \textcolor{orange}{$\rightarrow$~No observed abnormality} \textcolor{orange}{$\rightarrow$~No abnormalities noted} \textcolor{orange}{$\rightarrow$~No findings} \textcolor{orange}{$\rightarrow$~No notable abnormalities detected} \textcolor{orange}{$\rightarrow$~No clear abnormality identified} \textcolor{orange}{$\rightarrow$~No obvious cavities or masses detected} \textcolor{orange}{$\rightarrow$~No IMP Findings seen}

\textbf{Embedding clustering}\\
C-0: \textcolor{blue}{$\rightarrow$~No} \textcolor{cyan}{$\rightarrow$~no} \textcolor{cyan}{$\rightarrow$~no} \textcolor{cyan}{$\rightarrow$~No} \textcolor{cyan}{$\rightarrow$~No} \textcolor{orange}{$\rightarrow$~no} \textcolor{orange}{$\rightarrow$~no}\\
C-1: \textcolor{cyan}{$\rightarrow$~No evident abnormalities} \textcolor{cyan}{$\rightarrow$~No visible abnormalities} \textcolor{orange}{$\rightarrow$~No obvious abnormalities noted} \textcolor{orange}{$\rightarrow$~No abnormalities noted} \textcolor{orange}{$\rightarrow$~No notable abnormalities detected}\\
C-2: \textcolor{cyan}{$\rightarrow$~No, no evident focal opacities}\\
C-3: \textcolor{cyan}{$\rightarrow$~No evidenced abnormality}\\
C-4: \textcolor{cyan}{$\rightarrow$~No palpable evidence found}\\
C-5: \textcolor{cyan}{$\rightarrow$~No admireisible intraparenchymal abnormalities found.}\\
C-6: \textcolor{orange}{$\rightarrow$~No observed abnormality} \textcolor{orange}{$\rightarrow$~No clear abnormality identified}\\
C-7: \textcolor{orange}{$\rightarrow$~No findings}\\
C-8: \textcolor{orange}{$\rightarrow$~No obvious cavities or masses detected}\\
C-9: \textcolor{orange}{$\rightarrow$~No IMP Findings seen}

\textit{Observation:} NLI clustering consolidates semantically equivalent phrases into two cohesive groups, while embedding clustering fragments them into many small, form-based clusters, showing limited semantic integration.

\subsection{Example 2: "Lesion location" statements}

\noindent
\begin{tabular}{@{}m{0.72\textwidth} m{0.26\textwidth}@{}}
\raggedright
\textit{VQA Source:} VQA-RAD Test Set\\
\textit{Question:} What lobe of the brain is the lesion located in?\\
\textit{Reference answer:} Right frontal lobe\\
\textit{Answer variant requested:} SingleSentence\\
\textit{Model:} Qwen2.5-VL-7B-Instruct\\
&
\centering
\includegraphics[width=0.5\linewidth]{figures/images/radvqa_SingleSentence_984.png}
\end{tabular}\\

\textbf{NLI clustering}\\
C-0: \textcolor{blue}{$\rightarrow$~The lesion appears to be located in the left frontal lobe.} \textcolor{cyan}{$\rightarrow$~The lesion appears to be located in the left frontal lobe of the brain.} \textcolor{cyan}{$\rightarrow$~Left frontal lobe} \textcolor{cyan}{$\rightarrow$~Left frontal lobe.} \textcolor{orange}{$\rightarrow$~The lesion appears to be in the left frontal lobe.} \textcolor{orange}{$\rightarrow$~The lesion is located in the left frontal lobe.} \textcolor{orange}{$\rightarrow$~The lesion appears to be located in the left frontal lobe.}\\
C-1: \textcolor{cyan}{$\rightarrow$~Left parietal lobe.} \textcolor{cyan}{$\rightarrow$~The lesion appears to be located in the left parietal lobe.}\\
C-2: \textcolor{cyan}{$\rightarrow$~The lesion appears to be located in the left hemisphere, suggesting it could be in the frontal or parietal lobe.}\\
C-3: \textcolor{cyan}{$\rightarrow$~The lesion appears to be in the left hemisphere, but exact lobe localization is subjective and requires clinical correlation for definitive diagnosis.}\\
C-4: \textcolor{cyan}{$\rightarrow$~The lesion is localized in the right frontal lobe.} \textcolor{cyan}{$\rightarrow$~The lesion appears to be located predominantly in the right frontal lobe, with some involvement crossing to the left hemisphere.} \textcolor{orange}{$\rightarrow$~The lesion appears to be in the right frontal lobe.} \textcolor{orange}{$\rightarrow$~The lesion appears to be located in the right frontal lobe according to anatomical landmarks and imaging context.}\\
C-5: \textcolor{cyan}{$\rightarrow$~The lesion appears to be subcortical, not clearly in any specific lobe from this view.}\\
C-6: \textcolor{orange}{$\rightarrow$~Temporal lobe.}\\
C-7: \textcolor{orange}{$\rightarrow$~The lesion appears to be located in the left hemisphere, possibly the frontal or temporal region, but precise localization requires correlation with clinical symptoms.} \textcolor{orange}{$\rightarrow$~The lesion appears to be in the left frontal lobe of the brain.}\\
C-8: \textcolor{orange}{$\rightarrow$~The lesion appears to be in the parietal lobe.} \textcolor{orange}{$\rightarrow$~The lesion appears to be in the left parietal lobe.}

\textbf{Embedding clustering}\\
C-0: \textcolor{blue}{$\rightarrow$~The lesion appears to be located in the left frontal lobe.} \textcolor{cyan}{$\rightarrow$~Left parietal lobe.} \textcolor{cyan}{$\rightarrow$~The lesion appears to be located in the left frontal lobe of the brain.} \textcolor{cyan}{$\rightarrow$~The lesion appears to be located in the left hemisphere, suggesting it could be in the frontal or parietal lobe.} \textcolor{cyan}{$\rightarrow$~The lesion appears to be in the left hemisphere, but exact lobe localization is subjective and requires clinical correlation for definitive diagnosis.} \textcolor{cyan}{$\rightarrow$~The lesion is localized in the right frontal lobe.} \textcolor{cyan}{$\rightarrow$~Left frontal lobe.} \textcolor{cyan}{$\rightarrow$~The lesion appears to be located in the left parietal lobe.} \textcolor{cyan}{$\rightarrow$~The lesion appears to be located predominantly in the right frontal lobe, with some involvement crossing to the left hemisphere.} \textcolor{cyan}{$\rightarrow$~The lesion appears to be subcortical, not clearly in any specific lobe from this view.} \textcolor{cyan}{$\rightarrow$~Left frontal lobe.} \textcolor{orange}{$\rightarrow$~The lesion appears to be in the left frontal lobe.} \textcolor{orange}{$\rightarrow$~The lesion is located in the left frontal lobe.} \textcolor{orange}{$\rightarrow$~The lesion appears to be located in the left hemisphere, possibly the frontal or temporal region, but precise localization requires correlation with clinical symptoms.} \textcolor{orange}{$\rightarrow$~The lesion appears to be in the parietal lobe.} \textcolor{orange}{$\rightarrow$~The lesion appears to be in the left parietal lobe.} \textcolor{orange}{$\rightarrow$~The lesion appears to be in the right frontal lobe.} \textcolor{orange}{$\rightarrow$~The lesion appears to be located in the left frontal lobe.} \textcolor{orange}{$\rightarrow$~The lesion appears to be located in the right frontal lobe according to anatomical landmarks and imaging context.} \textcolor{orange}{$\rightarrow$~The lesion appears to be in the left frontal lobe of the brain.}\\
C-1: \textcolor{orange}{$\rightarrow$~Temporal lobe.}

\textit{Observation:} The NLI clustering forms clear anatomical groups (e.g., left/right, frontal/parietal), while the embedding clustering collapses most descriptions into a single large cluster, failing to distinguish laterality or lobe-level differences.

\section{Scalability Analysis of Computational Resources}
\label{app:scalability}

\begin{figure*}[htb]
    \includegraphics[width=1\textwidth]{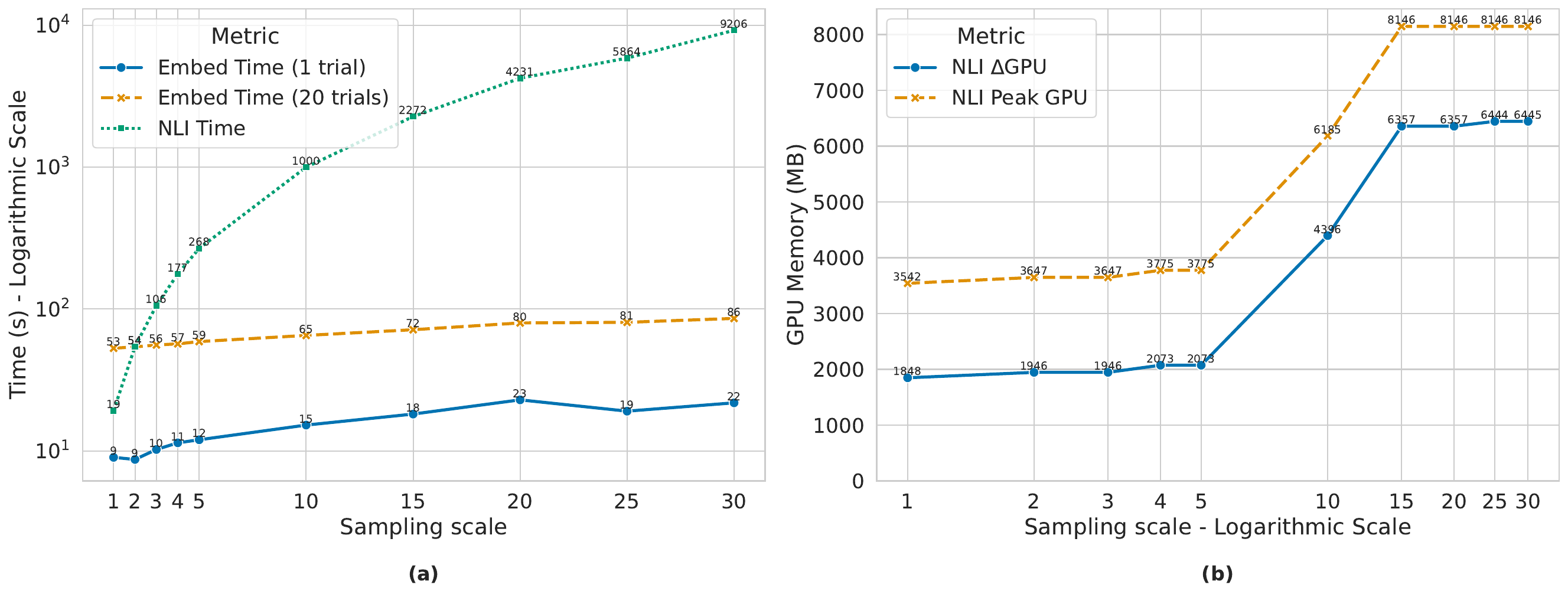}
    \centering
    \caption{Scalability analysis of computational resources for NLI-based and embedding-based hallucination detection, using Qwen2.5-VL-7B-Instruct on the VQA-RAD dataset. 
    (a) Processing time (logarithmic scale) and (b) NLI-related GPU memory usage, as a function of the maximum number of visual distortions (sampling scale). 
    While panel (a) shows the NLI-based approach scales heavily in time, panel (b) focuses only on its GPU cost. The embedding-based clustering is ultra-light in comparison, requiring only a stable ~160 MB $\Delta$GPU / ~1860 MB peak GPU across all distortion levels (not plotted).}
    \label{fig:scale_performance}
\end{figure*}

Figure~\ref{fig:scale_performance} illustrates how the computational requirements of the NLI-based and embedding-based clustering methods evolve across varying sampling scales.

\textbf{Processing time (Fig.~\ref{fig:scale_performance}a).}
The NLI-based curve rises sharply with increasing sampling scale, reflecting the rapid growth in the number of entailment relations that must be evaluated.  
The two embedding curves correspond to different threshold-selection strategies: a single-trial setting that represents the practical evaluation cost and a multi-trial setting that adds optional overhead from a more extensive search.  
Both embedding curves increase only gradually because, once embeddings are produced, the remaining similarity operations involve inexpensive pairwise computations.

\textbf{GPU memory usage (Fig.~\ref{fig:scale_performance}b).}
The memory plot reports two quantities.  
\emph{$\Delta$GPU} denotes the additional memory allocated during clustering relative to the allocation immediately before the operation, and  
\emph{peak GPU} indicates the maximum memory usage observed during execution.  
For the NLI-based method, both $\Delta$GPU and peak memory grow with sampling scale until the activation footprint of the MNLI batches (128 used here) saturates GPU usage, yielding the plateau observed at the largest scales.  
In contrast, the embedding-based method maintains a nearly flat memory profile across all sampling scales, as the embedding step dominates memory usage and subsequent similarity computations add only a small, scale-insensitive overhead.

\textbf{Overall implication.}
Together, the curves highlight a clear contrast in scalability: the NLI-based approach becomes increasingly demanding in both computation time and memory as sampling scales grow, whereas the embedding-based approach remains lightweight and stable.  
The single-trial embedding curve therefore reflects the practical runtime and memory envelope for scalable deployment of the method.

\end{document}